\documentclass[twoside,11pt]{article}

\usepackage{blindtext}

\usepackage{jmlr2e}

\usepackage[svgnames]{xcolor}
\usepackage{tikz}
\usepackage{subcaption}
\usepackage{amsmath}
\usepackage{algorithm2e}
\usepackage{mathtools}
\usepackage{booktabs}
\usepackage{multirow}
\usepackage{txfonts}

\usetikzlibrary{arrows.meta,arrows}
\tikzstyle{buyer} = [shape=rectangle, very thick, minimum width=13mm, rounded corners, draw, align=center]
\tikzstyle{seller} = [shape=rectangle, very thick, minimum width=13mm, rounded corners, draw, align=center]
\tikzstyle{caption} = [shape=rectangle, very thick, align=center]
\tikzstyle{computation} = [very thick,{Stealth[length=2mm, width=2mm]}-{Stealth[length=2mm, width=2mm]},>=to, RoyalBlue]
\tikzstyle{information} = [very thick, {Stealth[length=2mm, width=2mm]}-, Red]
\tikzstyle{monetary} = [very thick,-{Stealth[length=2mm, width=2mm]}, Green]
\tikzstyle{flow} = [very thick, -{Stealth[length=2mm, width=2mm]}]

\newtheorem{assumption}{Assumption}
\newtheorem{sublemma}{Lemma}[section]

\newcommand{\argmin}{\mathop{\mathrm{argmin}}}
\DeclareMathOperator*{\argmax}{argmax}

\usepackage{lastpage}
\jmlrheading{25}{2024}{1-\pageref{LastPage}}{10/23; Revised
5/24}{6/24}{23-1385}{Thomas Falconer, Jalal Kazempour and Pierre Pinson}

\ShortHeadings{Bayesian Regression Markets}{Falconer, Kazempour and Pinson}
\firstpageno{1}

\begin{document}

\title{Bayesian Regression Markets}

\author{\name Thomas Falconer \email falco@dtu.dk \\
       \addr Department of Wind and Energy Systems\\
       Technical University of Denmark\\
       Kgs. Lyngby, 2800, Denmark
       \AND
       \name Jalal Kazempour \email jalal@dtu.dk \\
       \addr Department of Wind and Energy Systems\\
       Technical University of Denmark\\
       Kgs. Lyngby, 2800, Denmark
       \AND
       \name Pierre Pinson \email p.pinson@imperial.ac.uk \\
       \addr Dyson School of Design Engineering\\
       Imperial College London\\
       London, SW7 2DB, United Kingdom}

\editor{Amos Storkey}

\maketitle

\begin{abstract}
    Although machine learning tasks are highly sensitive to the quality of input data, relevant datasets can often be challenging for firms to acquire, especially when held privately by a variety of owners. For instance, if these owners are competitors in a downstream market, they may be reluctant to share information.
    Focusing on supervised learning for regression tasks, we develop a \textit{regression market} to provide a monetary incentive for data sharing. Our mechanism adopts a Bayesian framework, allowing us to consider a more general class of regression tasks.
    We present a thorough exploration of the market properties, and show that similar proposals in literature expose the market agents to sizeable financial risks, which can be mitigated in our setup.
\end{abstract}

\begin{keywords}
  regression, bayesian inference, collaborative analytics, data markets, game theory
\end{keywords}

\section{Introduction}
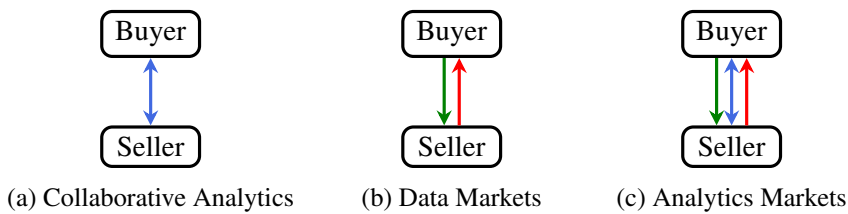
\begin{figure}[t]
    \centering
    \hspace{-5mm}
    \begin{subfigure}[]{0.3\textwidth}
        \centering
        \begin{tikzpicture}
            \node[buyer] (b0) at (0, 2.5) {Buyer};
            \node[seller] (s0) at (0, 1) {Seller};
            \draw[computation] (b0) -- (s0);
        \end{tikzpicture}
        \caption{Collaborative Analytics}
        \label{fig:collaborative_analytics}
    \end{subfigure}
    \hspace{0mm}
    \begin{subfigure}[]{0.2\textwidth}
        \centering
        \begin{tikzpicture}
            \node[buyer] (b1) at (0, 2.5) {Buyer}; 
            \node[seller] (s1) at (0, 1) {Seller}; 
            \begin{scope}[transform canvas={xshift=.1cm}]
                \draw[information] (b1) -- (s1);
            \end{scope}
            \begin{scope}[transform canvas={xshift=-.1cm}]
                \draw[monetary] (b1) -- (s1);
            \end{scope}
        \end{tikzpicture}
        \caption{Data Markets}
        \label{fig:data_markets}
    \end{subfigure}
    \hspace{1mm}
    \begin{subfigure}[]{0.25\textwidth}
        \centering
        \begin{tikzpicture}
            \node[buyer] (b2) at (0, 2.5) {Buyer}; 
            \node[seller] (s2) at (0, 1) {Seller}; 
            \draw[computation] (b2) -- (s2);
            \begin{scope}[transform canvas={xshift=.2cm}]
                \draw[information] (b2) -- (s2);
            \end{scope}
            \begin{scope}[transform canvas={xshift=-.2cm}]
                \draw[monetary] (b2) -- (s2);
            \end{scope}
        \end{tikzpicture}
        \caption{Analytics Markets}
        \label{fig:analytics_markets}
    \end{subfigure}
    \caption{Schematic illustration of existing frameworks for data sharing with multiple buyers and sellers, where each figure depicts a building block consisting of a single interaction. The blue, red and green arrows indicate computational, information and monetary transactions between the buyer and the seller, respectively.\protect\footnotemark}
  \label{fig:existing_framworks}
\end{figure}

As machine learning models continue to demand more data, practitioners often concentrate on challenges associated with data processing, feature selection and engineering, in addition to model building and validation, to optimize performance. These efforts are typically based on the assumption that data is readily available via some central authority, yet in practice, datasets are inherently distributed amongst owners with heterogeneous characteristics (e.g., privacy preferences).
This has motivated several developments in the field of collaborative analytics, also known as federated learning (Figure~\ref{fig:collaborative_analytics}), where models are trained on local servers without the need for data centralization, thereby preserving privacy and distributing the computational burden \citep{kairouz2021advances}.
However, this method for data sharing is \textit{incentive-free}, relying on the critical assumption that owners are willing to collaborate (i.e., by sharing their private information) altruistically. This strong assumption may be violated if owners are competitors in a downstream market environment \citep{galor1985information}. Consequently, a fruitful area of research has emerged that proposes to instead \textit{commoditize} data within a market-based framework, where compensation (e.g., remuneration) can be used as an incentive for collaboration \citep{bergemann2019markets}.

Information economics has been well established in game theory literature since the 1980s \citep{galor1985information}, with early works focusing on incentive-free data sharing, both publicly \citep{morris2002social} and within local information channels \citep{dahleh2016coordination}. Over recent years, interest in data monetization has grown rapidly, for which the first proposals considered \textit{data markets} (Figure~\ref{fig:data_markets}), allowing buyers to purchase raw data (i.e., features) from sellers through bilateral transactions \citep{rasouli2021data}.
Whilst a seemingly practical way to acquire data from others, the value of a feature to the buyer depends on the \textit{analytics task} at hand, hence pricing raw data in these markets is difficult \citep{cong2022pricing}, especially considering privacy-preservation \citep{acemoglu2022too}. Early game theoretic notions of informational efficiency in financial markets \citep{hayek1986use} also laid the groundwork for the closely related topic of \textit{prediction markets}, which are designed to aggregate information about the likelihood of future events.
These markets establish financial securities, with their payout contingent on the outcome of the events and trading prices construed as collective forecasts \citep{chen2010designing}. Traders effectively wager on repeatedly occurring \citep{bottazzi2019far} or one-shot \citep{manski2006interpreting} events.

\footnotetext{Adapted from \citet{pinson2022share}.}

Prediction markets are widely embraced within the machine learning community, with the common goal of forecasting future outcomes and familiar use of convex analysis \citep{frongillo2018bounded}. Recent works propose these markets as an effective means of crowdsourcing rows of data for machine learning by providing incentives to update a centralized hypothesis \citep{abernethy2011collaborative}.
So-called \textit{machine learning markets} extend this concept to complex multivariate prediction problems, where algorithms can compete using strategies that are either dynamic \citep{jahedpari2017online} or fixed, using betting functions \citep{barbu2012introduction, lay2012artificial}, utility functions \citep{storkey2011machine}, or static risk measures \citep{hu2014multi}, with the possibility of bounded regret for the owner of the task \citep{chen2010new, abernethy2013efficient}. The connection to data markets is clear, as both provide incentives for agents to disclose relevant private information. That being said, prediction markets are impractical for the general case of information sharing as support agents need to decide which tasks to make predictions for, yet the relevance of a dataset to a particular task is unknown \textit{a priori}, especially when considering the inevitability of overlapping information.

Instead, by recognizing that the commodity need not be data itself, ideas from both data markets and collaborative analytics can be combined to form \textit{analytics markets} (Figure~\ref{fig:analytics_markets}), real-time mechanisms that match features to analytics tasks based on the enhanced capabilities provided \citep{agarwal2019marketplace}.
As in \citet{pinson2022regression}, we focus on applications in which the analytics task describes a regression model along with the process for inference used for training (i.e., our attention centers on \textit{regression markets}). 
This builds upon current literature concerning data elicitation from strategic \citep{dekel2010regression} and privacy-sensitive \citep{cummings2015truthful} owners. In this context, owners of regression models seek to enhance predictive performance, for which they have a private valuation (e.g., their value of forecast accuracy in a downstream decision-making process).
Their public bids, which may not equal private valuations, are then used to set the price. Sellers propose their own data as features and are remunerated based on their marginal contributions to the improved model-fitting. The market revenue is therefore a function of both the market price and the overall enhancement in predictive performance.

In this work, we develop a regression market framework that enables Bayesian methods, allowing us to consider a more general class of regression tasks compared with the frequentist frameworks of previous works that treat parameters as fixed quantities \citep{pinson2022regression}.
Neglecting parameter estimation uncertainty can yield overly confident predictions that miss-represent the true level of variability in the data. In contrast, Bayesian analysis offers a principled framework for modelling parameter uncertainty that subsumes many frequentist regression methods, providing the buyer with richer and more nuanced information about future outcomes.
Treating parameters as random variables acknowledges that for a given regression task, the value of a particular feature, and hence the remuneration to its owner, is itself subject to uncertainty.
Conceptually, this is a paradigm shift in the field of regression markets and opens new avenues of research into uncertainty-aware mechanism designs for data sharing.

We also provide a thorough exploration of the market properties. In previous works (e.g., \citealp{agarwal2019marketplace}, \citealp{pinson2022regression}), such properties are only considered in expectation, however we demonstrate that many can be violated in a single shot of the market, exposing agents to considerable financial risks, especially when a limited number of observations are available.
We propose an alternative framework to value features based on the expected information gain provided as opposed to simply the impact on the objective value, which has the propensity to mitigate these risks entirely, despite being asymptotically equivalent.

The remainder of the paper is structured as follows: Section~\ref{sec:preliminaries} introduces the market agents and the design of our proposed market mechanism. 
Section~\ref{sec:market_properties} assesses the theoretical market properties of our proposal and presents methods for mitigating financial risks exhibited by the agents. 
Section~\ref{sec:simulation_studies} and Section~\ref{sec:real_world} illustrate our findings through a set of simulation-based and real-world case studies, respectively.
Finally, Section~\ref{sec:conclusion} gathers our conclusions and perspectives for future work.

\section{Market Setup} \label{sec:preliminaries}
We begin by presenting a general mechanism design for the analytics market, which is intended to be hosted on a platform capable of handling both the analytical (e.g., parameter inference) and market-based (e.g., revenue allocation) components together in tandem. 
The market comprises multiple agents---we define a \textit{transaction} as an exchange between a single \textit{central agent} (i.e., a buyer) and multiple \textit{support agents} (i.e., sellers), at a particular point in time, whereby the central agent seeks to enhance the predictive performance of a \textit{regression task}, for which the support agents propose their own data as input features.\footnote{Whilst this definition preserves the capacity for parallel transactions, we assume each is independent, thereby disregarding data exclusivity, wherein the same data can only be sold a finite number of times \citep{cao2017trading}, as well as any externalities this may exert \citep{bonatti2024selling}.}

Although there may be multiple sellers, the enhancement in performance received by the central agent is a function of the complete set of information available. We hence view a transaction as being between the central agent and a \textit{single} monopolistic support agent, a single agent with access to the complete set of features with only one item for sale---the available \textit{loss reduction}. 
The private valuation is assumed to equal the public bid (i.e., the central agent's valuation for a marginal improvement in model fitting).
The central agent is then allocated the full performance enhancement offered by the monopolistic support agent, and the payment collected is a function of these values. One can view this as a specification of the mechanism proposed in \citet{agarwal2019marketplace}, where the monopolistic support agent offers several possible performance enhancements, each representing varying degrees of obfuscation of the true data, characterized by the discrepancy between the bid of the central agent and the market price. 
Since we assume the market price is exogenous, our work is concerned specifically with the regression analysis and subsequent revenue allocation, as opposed to the pricing mechanism.

\subsection{Market Agents}
Let $\mathcal{A}$ denote the set of market agents, one of which $c \in \mathcal{A}$ is the central agent seeking to enhance their forecasts. The remaining agents $a \in \mathcal{A}_{-c}$ are support agents that propose data as input features, whereby $\mathcal{A}_{-c} = \mathcal{A} \setminus \{c\}$.
The central agent is characterized by their interest in a particular stochastic process $\{Y_t\}$, defined as a set of successive random variables $Y_t$ indexed over discretized time steps $t$. Eventually, a time-series $\{y_t\}$ is observed, comprising realizations from $\{Y_t\}$ (i.e., one per time step). Instead of assuming that a particular characteristic of $Y_t$ is sought (e.g., the expected value, a specific quantile, etc.), we rather model the entire distribution, albeit conditioned on the observed data; the characteristic extracted by the central agent is simply treated as some downstream decision-making process.

We write $\mathbf{x}_{\mathcal{I}, t}$ as the vector of input features at time $t$, indexed by the ordered set $\mathcal{I}$. Each agent $a \in \mathcal{A}$ owns a subset $\mathcal{I}_a \subseteq \mathcal{I}$ of indices, such that the features are distributed as follows: the central agent $c$ owns the subset $\mathcal{I}_c \subset \mathcal{I}$. Each support agent $a \in \mathcal{A}_{-c}$ also owns a subset of features, with indices $\mathcal{I}_a \subset \mathcal{I}$, such that $\vert\mathcal{I}_c\vert + \sum_{a \in \mathcal{A}_{-c}} \vert\mathcal{I}_a\vert = \vert\mathcal{I}\vert$. 
We write $\mathcal{I}_{-c}$ as the set of indices for features owned only by the support agents.
Since the data is observed at successive time steps, we let $\mathbf{x}_t = [x_{1, t}, \dots, x_{\vert\mathcal{I}\vert, t}]^{\top}$ be the vector of values for all features at time $t$. When only a particular subset of features $\mathcal{C} \subseteq \mathcal{I}$ is used, we add an index for the set itself, such that the vector of values for features in $\mathcal{C}$ at time $t$ is denoted by $\mathbf{x}_{\mathcal{C}, t}$. We write $\mathcal{D}_{\mathcal{C}, t} = \{\mathbf{x}_{\mathcal{C}, t^{\prime}}, y_{t^{\prime}}\}_{\forall t^{\prime} \leq t}$ to be the set of input-output pairs for a particular subset of features observed over a set of discrete time indices $t^{\prime} \in \{1, \dots, t\}$ up until time $t$.

\subsection{Regression Task} 
To instigate a transaction, the central agent first posts a regression task to the market platform, which describes the particular model for which they seek to enhance predictive performance. We consider the problem of interpolating through data (i.e., the observations $\{y_t\}$) under the assumption that the target signal is subject to noise, whilst the input features are noise-free. 
Let us define an interpolant as a mapping $f$ between a subset of features $\mathbf{x}_{\mathcal{C}, t}$ and a real-valued scalar, which may represent the expected value of the target signal conditioned on the inputs such that
\begin{equation}
    f : \mathbf{x}_{\mathcal{C}, t} \in \mathbb{R}^{\vert \mathcal{C} \vert} \mapsto \mathbb{E} [ Y_t \, \vert \, \mathbf{x}_{\mathcal{C}, t} ] \in \mathbb{R}, \quad \forall t, \ \forall \mathcal{C}.
    \label{eq:regression_mapping}
\end{equation}
We focus solely on parametric regression, and further limit ourselves to functions that can be expressed as linear in their coefficients, with a view to preserve convexity and later guarantee certain market properties. We obtain a rich class of models by considering linear combinations of a fixed set of nonlinear functions (i.e., basis functions). 
Let $\mathbf{w}_{\mathcal{C}} \in \mathbb{R}^{\vert \mathcal{C} \vert}$ be a vector of coefficients that is used to parameterize the mapping in (\ref{eq:regression_mapping}), which, for notational brevity, we assume to be part of a general set of free parameters $\Theta_{\mathcal{C}}$ that shall be inferred from data. We write $\varphi(\mathbf{x}_{\mathcal{C}, t})$ to be the vector of basis functions specified by the central agent, such that the linear interpolant is given by
\begin{equation}
    f(\mathbf{x}_{\mathcal{C}, t}, \mathbf{w}_{\mathcal{C}}) = \mathbf{w}_{\mathcal{C}}^\top \, \varphi(\mathbf{x}_{\mathcal{C}, t}), \quad \forall t, \ \forall \mathcal{C},
    \label{eq:interpolated_function}
\end{equation}
where we assume that the vector of basis functions under consideration invariably incorporates a dummy basis function (i.e., $\varphi_0(\mathbf{x}_{\mathcal{C}, t}) = 1, \, \forall t$) which is included as part of the feature set owned by the central agent. 
\begin{remark}
    In general, the central agent need not own any feature themselves and thereby crowd-source their predictions, in which case only the dummy term is provided and all predictive performance is supplied by the features owned by support agents.
\end{remark}

We model the target variable as a deviation from the deterministic mapping in (\ref{eq:interpolated_function}) under a zero-mean additive noise process, the parameters of which are also held in $\Theta_{\mathcal{C}}$.
Bayesian inference treats the parameters as random variables and aims to infer their distribution by incorporating prior beliefs, which are updated as new data is observed. Let $h \in \mathcal{H}$ be a hypothesis, a set of fixed assumptions that restricts the space of possible regression models, comprising the vector of basis functions, as well as the functional forms of two probability distributions: both the prior (i.e., plausible parameter values) and the likelihood (i.e., the probability of the data conditioned on the parameters). The regression task posted to the market platform by the central agent at time $t$ is therefore fully described by a hypothesis and the observed data.


\subsection{Market Clearing} \label{sec:market_clearing}
We suppose each support agent is willing to accept any nonnegative payment if their data is deemed useful. However, we acknowledge that support agents may prefer to condition their participation on a minimum payment to, for instance, reflect privacy costs \citep{acquisti2016economics}. 
Certain features in the market may also be irrelevant for the regression task, hence we assume the following.

\begin{assumption} \label{as:feature_selection}
    Given the specified hypothesis, the market operator is tasked with selecting relevant features (e.g., by means of cross-validation), such that only those that reduce the expected value of the loss function are considered.
\end{assumption}

As our problem is convex, any additional feature cannot increase the in-sample loss in expectation. Therefore we refer here to the out-of-sample loss, evaluated for instance with cross-validation, since by submitting a feature to the market a support agent provides the market operator with training data. 
For discussions on conventional feature selection problems, cross-validation in a Bayesian context and methods for marginal likelihood optimization, the reader is referred to
\citet{guyon2003introduction}, \citet{watanabe2010asymptotic} and  \citet{fong2020marginal}, respectively.
\begin{remark}
    Adopting an exogenous feature selection process prior to market clearing is merely to align with standard practices in machine learning operations (MLOps). One could also consider feature selection processes that are endogenous or that occur after the market has been cleared, by employing informative priors \citep{han2022lasso} or applying post-processing steps \citep{liu2020absolute}, respectively, albeit at a cost to the market. In particular, these approaches may undermine potential improvements in predictive performance or lead to skewed payments as a price to pay for avoiding such independent feature selection.
\end{remark}

Once the entire set of required market inputs have been received, the market operator is tasked with clearing the market. This procedure involves several steps, namely parameter inference, performance evaluation, payment collection and revenue allocation.

\subsubsection{Parameter Inference}
Based on all of the observations up until time $t$, we can summarize our updated beliefs regarding the parameters through the posterior distribution, which, by virtue of Bayes theorem, is proportional to the product of the likelihood and the prior such that
\begin{equation}
    p(\Theta_{\mathcal{C}} \vert \mathcal{D}_{\mathcal{C}, t}) \propto p( \mathcal{D}_{\mathcal{C}, t} \vert \Theta_{\mathcal{C}}) p(\Theta_{\mathcal{C}}), \quad \forall t, \ \forall \mathcal{C}. \label{eq:bayes_theorem} 
\end{equation}

For an arbitrary choice of prior, the posterior may not be available in closed-form, necessitating methods for approximate Bayesian inference (e.g., Monte-Carlo integration) to be employed. However, for a known functional form of the likelihood, priors that are conjugate can result in posteriors with tractable, well-known densities.
It may be appropriate to allow the moments of this distribution to vary in time, thereby accounting for nonstationarities in any of the underlying processes that can lead to concept drift. 
In a Bayesian treatment of linear regression, batch inference can be viewed as a specification of this more general \textit{online learning} problem, whereby the parameters are updated in a recursive manner. To see this, we re-write the expression in (\ref{eq:bayes_theorem}) as a series of sequential updates such that
\begin{subequations}
    \begin{alignat}{2}
        p(\Theta_{\mathcal{C}} \vert \mathcal{D}_{\mathcal{C}, t}) &\propto p( \mathcal{D}_{\mathcal{C}, t} \vert \Theta_{\mathcal{C}}) p(\Theta_{\mathcal{C}} \vert  \mathcal{D}_{\mathcal{C}, t-1}), &&\quad \forall t, \ \forall \mathcal{C}, \label{eq:bayes_theorem_online} \\
        &= p( \mathcal{D}_{\mathcal{C}, t} \vert \Theta_{\mathcal{C}}) \left[ p(\Theta_{\mathcal{C}}) \prod_{t^\prime < t} p( \mathcal{D}_{\mathcal{C}, t^\prime} \vert \Theta_{\mathcal{C}}) \right], &&\quad \forall t, \ \forall \mathcal{C}. \label{eq:sequential_updates}
    \end{alignat}
\end{subequations}

To place greater weight on more recent data, we can augment this update step to use exponential forgetting, where the importance given to past information decreases exponentially. This generally translates to the idea of likelihood flattening, whereby we reformulate (\ref{eq:sequential_updates}) as a trade-off between the posterior at the previous time step and the original prior (i.e., before any data had been observed), thereby emulating a loss in belief with respect to the historic estimates \citep{peterka1981bayesian}. This trade-off between the two distributions can be framed as the problem of finding the probability density function with minimum expected Kullback–Leibler (KL) divergence (i.e., relative entropy) between them \citep{kulhavy1993general}, which has a unique solution enabling us to replace the prior at time $t$ in (\ref{eq:bayes_theorem_online}) with the following:
\begin{subequations}
    \begin{alignat}{2}
        p(\Theta_{\mathcal{C}} \vert \mathcal{D}_{\mathcal{C}, t-1}, \tau) 
        &=\argmin_{p^\ast} \ \tau \, D_{\textrm{KL}} \left( p^\ast \, \vert\vert \, p(\Theta_{\mathcal{C}} \, \vert \mathcal{D}_{\mathcal{C}, t-1}) \right) + (1 - \tau) \, D_{\textrm{KL}} \left( p^\ast \,  \vert \vert \,  p(\Theta_{\mathcal{C}})  \right), \quad &&\forall t, \ \forall \mathcal{C}, \\
        &\propto p(\Theta_{\mathcal{C}} \vert \mathcal{D}_{\mathcal{C}, t-1})^{\tau} p(\Theta_{\mathcal{C}})^{1-\tau}, \quad &&\forall t, \ \forall \mathcal{C},
        \label{eq:posterior_flattening}
    \end{alignat}
\end{subequations}
where the variable $p^\ast$ denotes the resultant density function, $D_{\textrm{KL}}(\, \cdot \, \vert\vert \, \cdot \,) \in \mathbb{R}_{+}$ is the KL divergence and the parameter $\tau \in [0, 1]$ is analogous to the forgetting factor in time-weighted Least-Squares fitting \citep{vahidi2005recursive}.
Observe that, as $\tau \mapsto 1$, the prior information available at time $t$ becomes identical to the posterior information at the previous time step as in (\ref{eq:sequential_updates}), emulating batch learning, whereas when $\tau=0$, the previous information is \textit{forgotten} and we resort to the original (i.e., flat) prior. For convenience, we treat $\tau$ as a time-invariant hyperparameter, however for a full Bayesian treatment one could also infer its value jointly, together with $\Theta_{\mathcal{C}}$.

\subsubsection{Performance Evaluation}
Given observations up until time $t$, we can evaluate the performance of a specific subset of features by making a prediction for a time step $t^\ast$, conditioned on the observed input features. For now, we consider the general case where $t^\ast$ is an arbitrary time step to account for both in-sample (i.e., $t^\ast \leq t$) and out-of-sample (i.e., $t^\ast > t$) situations. In Bayesian regression analyses, a \textit{prediction} is typically defined to be the computation of the posterior predictive distribution, derived by integrating out the parameters using the convolution of the likelihood with the posterior, given by
\begin{align}
    p(y_{t^\ast} \vert \textbf{x}_{\mathcal{C}, t^\ast}, \mathcal{D}_{\mathcal{C}, t}) = \int p(y_{t^\ast} \vert \textbf{x}_{\mathcal{C}, t^\ast}, \mathcal{D}_{\mathcal{C}, t}, \Theta_{\mathcal{C}}) p(\Theta_{\mathcal{C}} \vert \mathcal{D}_{\mathcal{C}, t}) d \Theta_{\mathcal{C}}, \quad \forall \mathcal{C},
    \label{eq:predictive_distribution}
\end{align}
which for brevity we hereafter omit the training dataset and write as $p(y_{t^\ast} \vert \textbf{x}_{\mathcal{C}, t^\ast})$.
In order to evaluate predictive performance, we define a loss function $\ell$.
If a model describing a particular characteristic of $Y_t$ is sought, then this loss function could be set as a direct function of the residuals (i.e., by extracting the corresponding point from the predictive distribution).
However, as we intend to provide the entire predictive distribution, we can generally define $\ell$ as a function of the predictive density (i.e., $\ell_{\mathcal{C}, t^\ast} : p(y_{t^\ast} \vert \textbf{x}_{\mathcal{C}, t^\ast}) \mapsto \mathbb{R}$), assuming the following. 
\begin{assumption} \label{as:scoring_rule} The mapping $\ell$ is a  negatively-oriented strictly proper scoring rule. Accordingly, it holds that: (i) for any two models, the one that provides the more accurate description of the data will render a lower score; and (ii) the lowest score is uniquely obtained when the prediction converges to the true distribution.
\end{assumption}

In an online setup with exponential forgetting, evaluating $\ell$ at each time step can be perceived as a recursive and adaptive time-varying estimator of its expected value; adaptive in the sense that a greater weight is placed on more recent data. Hence, the in-sample estimate of $\mathbb{E}[\ell]$ for a particular subset of features at time $t$ with respect to (\ref{eq:predictive_distribution}) can be described by the following recursion:
\begin{align}
\mathbb{E}[\ell_{\mathcal{C}}]_{t} = (1 - \tau) \,  \ell_{\mathcal{C}, t} + \tau \, \mathbb{E}[\ell_{\mathcal{C}}]_{t-1}, \quad &\forall t, \ \forall \mathcal{C},
    \label{eq:time_varying_loss}
\end{align}
where to consider the case of out-of-sample evaluation (e.g., if $t$ is the next available time step) we simply replace $\mathcal{D}_{\mathcal{C}, t}$ in (\ref{eq:time_varying_loss}) with the most recent set of observations, $\mathcal{D}_{\mathcal{C}, t-1}$. 

\subsubsection{Payment Collection}
As well as a regression task, our market requires the central agent to post to the platform their public bid, denoted by $\lambda \in \mathbb{R}_{+}$, which represents an exogenous linear mapping between a unit improvement in $\ell$ and the corresponding downstream monetary reward that would be earned, thereby determining the market price.
We indeed acknowledge the weakness of this linearity assumption, as in practice, $\lambda$ may be, for instance, a logarithmic function of the central agent's revenue (i.e., further reductions in $\ell$ may provide diminishing returns), albeit with exponential costs for the support agents.
Nevertheless, we leave it as future work to explore the optimal functional form of $\lambda$. 
The market revenue at time $t$ is equal to the payment collected from the central agent, denoted $\pi_{c, t}$, which is a function of $\lambda$, as well as the overall improvement in the objective, such that
\begin{align}
    \pi_{c, t} = \lambda \, \left( \, \mathbb{E}[\ell_{\mathcal{I}_c}]_{t} - \mathbb{E}[\ell_{\mathcal{I}}]_{t} \right), \quad \forall t.
    \label{eq:buyer_payment}
\end{align}

\subsubsection{Revenue Allocation} \label{sec:revenue_allocation}
Once the market has been cleared, the natural question that follows is: \textit{how can we fairly allocate market revenue amongst support agents?} To answer this question, several auction-based setups have been proposed, considering topics such as privacy \citep{koutsopoulos2015auction}, data exclusivity \citep{cao2017trading} and negative externalities exhibited by the market agents \citep{bonatti2024selling}. Other methods bear upon interoperability in machine learning, using widely adopted solution concepts (namely, semivalues) for the problem of attribution in cooperative game theory to allocate revenue amongst support agents directly \citep{dubey1981value}. The benefit of this approach being that these solution concepts are generally characterized by a collection of axioms that yield desirable market properties by design \citep{ghorbani2019shapley}, specifically: symmetry, efficiency, null-player and additivity.
For a definition of these axioms, the reader is referred to \citet{chalkiadakis2011computational}.

If we frame features as players and their interactions as a cooperative game, the semivalue of a feature can be defined as its expected marginal contribution towards a set of other features, weighted solely based on the size of the sets. For many applications, the semivalue of choice is the \textit{Shapley value} \citep{shapley1997value}, the unique value that satisfies all of the four axioms stated above. Given the set $\mathcal{I}_{-c}$ of indices corresponding to features owned by the support agents, let $v : \mathcal{C} \in \mathcal{P}({\mathcal{I}_{-c}}) \mapsto \mathbb{R}$ be a characteristic function that maps the power set $\mathcal{P}(\mathcal{I}_{-c})$ of all features with indices in $\mathcal{I}_{-c}$ to a real-valued scalar, where the set $\mathcal{C}$ denotes a coalition in the
cooperative game. The Shapley value is given by
\begin{align}
    \phi_{i, t} = \sum_{\mathcal{C} \in 
    \mathcal{P}(\mathcal{I}_{-c} \setminus \{i\})
    } \frac{\vert\mathcal{C}\vert!(\vert\mathcal{I}_{-c}\vert - \vert\mathcal{C}\vert - 1)!}{\vert\mathcal{I}_{-c}\vert!} \, m_{i, t}(\mathcal{C} \cup \{\mathcal{I}_c\}), \quad \forall i \in \mathcal{I}_{-c}, \, \forall t,
    \label{eq:shapley_value}
\end{align}
where $m_{i, t}(\cdot)$ is the marginal contribution, often defined as $m_{i, t}(\cdot) = v_t (\cdot) - v_t (\cdot \cup \{i\})$ in relation to the characteristic function.\footnote{The weight in this discrete expectation assigned to each coalition is defined as such to avoid unnecessary calculations of the marginal contribution of the $i$-th feature to permutations of the same coalition, which would have equal value by virtue of the symmetry axiom. For instance, $v_{t}(\{i, j\}) \equiv v_{t}(\{j, i\}), \, \forall (i, j) \in \mathcal{I}_{-c}, \ i \neq j$, thus it is computationally favourable to avoid making this calculation twice.}

The Shapley value is then used to allocate market revenue. Given our estimator of the expected loss varies with time, attributions are time-varying too, as well as the market revenue in (\ref{eq:buyer_payment}). In line with (\ref{eq:time_varying_loss}), the expected Shapley value at time $t$ is given by
\begin{align}
    \mathbb{E}[\phi_i]_{t} = (1 - \tau) \phi_{i, t} + \tau \mathbb{E}[\phi_i]_{t-1}, \quad \forall i \in \mathcal{I}_{-c}, \, \forall t.
    \label{eq:recursive_shapley}
\end{align}

Then, as we evaluate (\ref{eq:recursive_shapley}) for each feature, the overall payment received by each support agent is simply given by
\begin{align}
    \pi_{a, t} = \sum_{i \in \mathcal{I}_a} \lambda \, \mathbb{E}[\phi_i]_{t} , \quad \forall a \in \mathcal{A}_{-c}, \ \forall t.
    \label{eq:seller_payment}
\end{align}

\subsection{Market Dynamics}
In practice, MLOps pipelines are typically divided into in-sample (i.e., training) and out-of-sample (i.e., testing) stages. The first stage involves Bayesian inference using observed input-output pairs, whilst at
the second stage, a trained model is used for genuine forecasting on previously unseen data, testing its capacity to generalize beyond the training set.
We hence adopt the two-stage regression market model proposed in \citet{pinson2022regression}, that is, the value of a feature is assessed based on marginal contributions to both the in-sample and out-of-sample estimates of $\mathbb{E}[ \ell ]$, albeit in separate transactions.

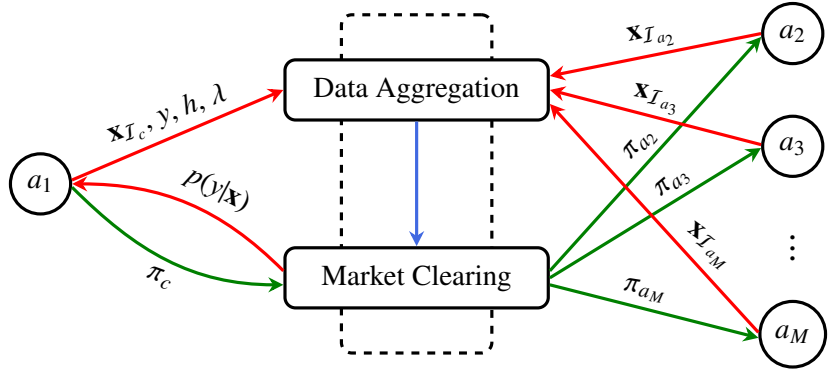
\begin{figure}[t]
    \centering
    \begin{tikzpicture}
            \node[shape=rectangle, rounded corners, draw, align=center, minimum height=45mm, minimum width=20mm, dashed, very thick] (market0) at (0, 0) {};
            \node[shape=rectangle, very thick, rounded corners, draw, align=center, minimum width=35mm, minimum height=8mm, fill=white] (market0) at (0, 1.25) {Data Aggregation};
            \node[shape=rectangle, very thick, rounded corners, draw, align=center, minimum width=35mm, minimum height=8mm, fill=white] (market2) at (0, -1.25) {Market Clearing};
            \node[shape=circle, very thick, rounded corners, draw, align=center, minimum width=8mm, minimum height=8mm] (central) at (-5, 0) {$a_1$};
            \node[shape=circle, very thick, rounded corners, draw, align=center, minimum width=8mm, minimum height=8mm] (s0) at (5, 2) {$a_2$};
            \node[shape=circle, very thick, rounded corners, draw, align=center, minimum width=8mm, minimum height=8mm] (s1) at (5, 0.5) {$a_3$};
            \node[shape=rectangle] (s2) at (5, -0.75) {\Large $\vdots$};
            \node[shape=circle, very thick, rounded corners, draw, align=center, minimum width=8mm, minimum height=8mm] (s2) at (5, -2) {$a_M$};
            
            \draw[flow, color=RoyalBlue] (market0.south) to (market2.north) node[black,midway, above, rotate=20] {};
            
            \draw[flow, color=Red] (central.7) to (market0.180) node[black,midway, above, xshift=-32mm, yshift=6mm, rotate=20] {$\textbf{x}_{\mathcal{I}_c}$, $y$, $h$, $\lambda$};
            \draw[flow, color=Green] (central.-6) to[out=-45,in=180] (market2.183) node[black,midway, above, xshift=-28mm, yshift=-3.5mm, rotate=-28] {$p(y \vert \textbf{x})$};
            \draw[flow, color=Red] (market2.177) to[out=135,in=0] (central.0) node[black,midway, above, xshift=-35mm, yshift=-15mm, rotate=-20] {$\pi_c$};
            
            \draw[flow, color=Red] (s0.180) to (market0.6) node[black, midway, above, xshift=32mm, yshift=16.5mm, rotate=12] {$\textbf{x}_{\mathcal{I}_{a_2}}$};
            \draw[flow, color=Green] (market2.3) to (s0.186) node[black, midway, above, xshift=32mm, yshift=3.5mm, rotate=50] {$\pi_{a_2}$};
            
            \draw[flow, color=Red] (s1.177) to (market0.0) node[black, midway, above, xshift=31.5mm, yshift=8mm, rotate=-9] {$\textbf{x}_{\mathcal{I}_{a_3}}$};
            \draw[flow, color=Green] (market2.0) to (s1.183) node[black, midway, above, xshift=36mm, yshift=-2mm, rotate=40] {$\pi_{a_3}$};
            
            \draw[flow, color=Red] (s2.177) to (market0.-6) node[black, midway, above, xshift=37mm, yshift=-11mm, rotate=-36] {$\textbf{x}_{\mathcal{I}_{a_{M}}}$};
            \draw[flow, color=Green] (market2.-3) to (s2.186) node[black, midway, above, xshift=30mm, yshift=-17mm, rotate=-9] {$\pi_{a_M}$};
        \end{tikzpicture}
    \caption{Overview of in-sample market platform operations at time $t$, with agent $a_1$ as the central agent and the total number of support agents $M = \vert \mathcal{A}_{-c} \vert$. The time index $t$ is omitted for brevity. Recall that the blue, red and green arrows indicate computational, information and monetary transactions, respectively.}
    \label{fig:market_overview}
\end{figure}

We round off this section with a summary of the key market platform operations---see Figure~\ref{fig:market_overview} for a graphical overview. The steps in Algorithm~\ref{alg:market_dynamics} occur until termination (i.e., when a specified time step $T$ has been reached). The out-of-sample market operations are similar, the only differences being that the posterior is not updated before predictions are made, and only when the observation of the target signal arrives is the revenue collected and allocated.

\RestyleAlgo{ruled}
\SetKwComment{Comment}{/* }{ */}
\begin{algorithm}
\caption{In-sample online regression market}\label{alg:two}
    \textbf{Input:} $\textbf{x}_{\mathcal{I}_c, t} \in \mathbb{R}^{\vert \mathcal{I}_c \vert}$, $y_{t} \in \mathbb{R}$, $\lambda \in \mathbb{R}_+$, $h \in \mathcal{H}$ \\
    \textbf{Output:} $p(y_t \vert \textbf{x}_t)$, $[\pi_{a, t} : a \in \mathcal{A}_{-c}]$ \\

    \While{$t \leq T$}{
        $p(\Theta_{} \vert \mathcal{D}_{t}, \tau) =  p( \mathcal{D}_{t} \vert \Theta_{\mathcal{C}}) p(\Theta \vert \mathcal{D}_{t-1}, \tau)$ \Comment*[r]{Update posterior}
        $p(y_{t} \vert \textbf{x}_{t}) = \int p(y_{t} \vert \textbf{x}_{t}, \Theta) p(\Theta_{} \vert \mathcal{D}_{t}) d \Theta$ \\
        $\pi_{c, t} = \lambda (\mathbb{E}[\ell_{\mathcal{I}_c}]_{t} - \mathbb{E} [\ell_{\mathcal{I}}]_{t})$ \Comment*[r]{Revenue extracted from central agent}
        \For{$a \in \mathcal{A}_{-c}$}{
            \For{$i \in \mathcal{I}_a$}{
                $\mathbb{E}[\phi_i]_{t} = (1 - \tau)\sum_{\mathcal{C} \in \mathcal{P}(\mathcal{I}_{-c} \setminus \{i\})} \frac{\vert\mathcal{C}\vert!(\vert\mathcal{I}_{-c}\vert - \vert\mathcal{C}\vert - 1)!}{\vert\mathcal{I}_{-c}\vert!} \, m_{i, t}(\mathcal{C} \cup \mathcal{I}_{c}) + \tau \mathbb{E}[\phi_i]_{t-1}$ \\
                $\pi_{a, t} = \pi_{a, t} + \lambda \mathbb{E}[\phi_i]_{t}$ \Comment*[r]{Revenue allocated to support agents}
            }
        }   
    } 
    \label{alg:market_dynamics}
\end{algorithm}

\begin{remark}  
    In our proposed mechanism, the market price is solely dependent on the valuation of the central agent, $\lambda$, demonstrating that the value of data need not be an intrinsic property, but rather derived from its contribution to the particular analytics task at hand.
    This indeed assumes features are interchangeable provided the same inferences can be drawn from either, but moreover, the inevitable overlapping information between features would render parameterized valuations for each subset computationally intractable.
\end{remark}

\section{Market Properties} \label{sec:market_properties}
The remaining design decision that will affect the market properties relates to the choice of characteristic function, $v_t(\cdot)$, used to value a coalition of features. The Shapley value has indeed emerged as the \textit{de facto} tool for interpreting predictions from complex machine learning models \citep{strumbelj2010efficient, sundararajan2020many, tsai2023faith}, yet its application in probabilistic contexts is not yet as well studied, as comparing models outputs as probability distributions is less straightforward that scalars. 
In general, at a given time $t$ we can set the characteristic function to be equal to the current estimate of the expected loss, described in (\ref{eq:time_varying_loss}), such that $v_t(\mathcal{C}) = \mathbb{E}[\ell_{\mathcal{C}}]_{t}$, hence the design decision is not the characteristic function itself \textit{per se}, but the particular functional form of $\ell$, which recall maps the predictive density to a real value.

In this section we introduce the following market designs: (i) $\mathcal{M}^{\mathrm{MLE}}_{\mathrm{NLL}}$---a frequentist framework based on maximum likelihood estimation (MLE) which values features using the negative logarithm of the likelihood (NLL), (ii) $\mathcal{M}^{\mathrm{BLR}}_{\mathrm{NLL}}$---the analogue of $\mathcal{M}^{\mathrm{MLE}}_{\mathrm{NLL}}$ now in a Bayesian linear regression (BLR) framework, and (iii) $\mathcal{M}^{\mathrm{BLR}}_{\mathrm{KL}-m}$ and $\mathcal{M}^{\mathrm{BLR}}_{\mathrm{KL}-v}$---BLR frameworks that instead value features based on the information gain they provide, measured using the KL divergence.

\subsection{Likelihood-based Designs}
In \citet{pinson2022regression}, a maximum likelihood framework is adopted, treating parameters as fixed, albeit unknown, quantities. The characteristic function can then simply be set to the expected value of the NLL.
We denote this frequentist market design by $\mathcal{M}^{\mathrm{MLE}}_{\mathrm{NLL}}$. In the following, we shall analyze the market properties obtained by extending this idea to its Bayesian analogue. 

In Bayesian regression we have access to the posterior distribution, from which revenue allocations derived using any random sample could be considered plausible with respect to the frequentist design. To attain the most nuanced representation of uncertainty, we instead provide the predictive distribution derived by marginalizing over the entire space of parameters. A reasonable candidate for the characteristic function is therefore again the NLL, which now incorporates the uncertainty in the parameter estimates in (\ref{eq:predictive_distribution}), such that
\begin{align}
    \ell_{\mathcal{C}, t} = - \log (p(y_{t} \, \vert \, \textbf{x}_{\mathcal{C}, t})), \quad &\forall t, \ \forall \mathcal{C},
    \label{eq:log_likelihood}
\end{align}
with $\mathcal{M}^{\mathrm{BLR}}_{\mathrm{NLL}}$ denoting the corresponding market design. In order for Assumption~\ref{as:scoring_rule} to be satisfied, the predictive density must be log-concave. Whilst many common distributions are indeed log-concave, and hence could be utilized easily in a maximum likelihood framework, in order to avoid approximation errors in general Bayesian inference we require the posterior to be available in closed-form, thus the prior and posterior should be conjugate. Therefore, as well as for mathematical convenience, to adhere to this we further assume the following, and leave exploration of alternative hypotheses to future work.
\begin{assumption} \label{as:gaussian_hypothesis} The hypothesis space $\mathcal{H}$ comprises only Gaussian likelihood functions along with a conjugate uninformative Gaussian prior.
\end{assumption}
\begin{remark}
    Whilst Assumption~\ref{as:gaussian_hypothesis} is restrictive, it is in fact common in practice (i.e., it is a byproduct of simply using mean-squared error in frequentist regression methods), and still permits non-Gaussian data generating processes, but merely induces misspecifiations in such a case.
\end{remark}

It is worth highlighting the tangible benefits to the central agent of transitioning from frequentist to Bayesian regression analyses. For instance, the additional element of predictive uncertainty provides richer and more nuanced information about future outcomes. In addition, maximum likelihood estimation also has a tendency to render implausible overparameterized models that generalize poorly to out-of-sample analyses. This is especially true when the number of training observations is limited, since increasing model complexity inevitably results in overfitting.
In contrast, Bayesian methods inherently embody \textit{Occam's razor} (i.e., a  proclivity towards simplicity) by exploiting prior knowledge that induces regularization without the need for ad-hoc penalty terms, thereby facilitating well-calibrated uncertainty estimates using training data alone.

We now explore the key properties of the likelihood-based Bayesian regression market. These properties are derived from the axioms that characterize the semivalue, all four of which are satisfied by the Shapley value. We first present the properties that we refer to as \textit{universal}, those which are guaranteed  to be satisfied under all circumstances. 
\begin{theorem} \label{the:universal_properties} Likelihood-based Bayesian regression markets of this kind yield the following universal market properties:
\begin{enumerate}
    \item Symmetry---two features $x_{i, t}$ and $x_{j, t}$ with equal marginal contribution to any coalition receive the same attribution, that is, $\forall \mathcal{C} \in \mathcal{I}_{-c} \setminus \{i, j\} : v_t (\mathcal{C} \cup \mathcal{I}_c \cup \{i\}) \equiv v_t (\mathcal{C} \cup \mathcal{I}_c \cup \{j\}) \mapsto \phi_{i, t} \equiv \phi_{j, t}, \ \forall (i, j) \in \mathcal{I}_{-c}, \ i \neq j, \ \forall t$.
    \item Linearity---for any two features $x_{i, t}$ and $x_{j, t}$, their joint contribution to a particular coalition of other features is equal to the sum of their marginal contribution, that is, $v_t (\mathcal{C} \cup \mathcal{I}_c \cup \{i\}) + v_t (\mathcal{C} \cup \mathcal{I}_c \cup \{j\}) = v_t (\mathcal{C} \cup \mathcal{I}_c \cup \{i, j\}), \ \forall \mathcal{C} \in \mathcal{I}_{-c} \setminus \{i, j\}, \ \forall t$.
    \item Budget balance---the payment of the central agent is equal to the sum of revenues received by the support agents, that is, $\pi_{c, t} \equiv \sum_{a \in \mathcal{A}_{-c}} \pi_{a, t}, \ \forall t$.
\end{enumerate}
\noindent {\bf Proof} Omitted since each universal property follows directly from the semivalue axioms satisfied by the Shapley value.
\hfill\BlackBox
\end{theorem}

With symmetry, attributions are invariant to permutation of indices, equivalent to the anonymity property in \citet{lambert2008self}, whilst linearity ensures that revenues remain consistent regardless of whether the features are offered individually or as a bundle, removing any incentive to strategically package features. 
Budget balance is a byproduct of efficiency, which states that total attribution allocated to all features should sum to the value of the grand coalition, that is, $v_t (\mathcal{I}) = \sum_{i \in \mathcal{I}_{-c}} \phi_{i, t}, \ \forall t$. Accordingly, given the definitions in (\ref{eq:buyer_payment}) and (\ref{eq:seller_payment}), it holds universally that the total sum of the revenues of each support agent equals the payment collected from the central agent.

In addition to these universally held market properties, our likelihood-based Bayesian regression market further obtains a collection of properties that we hereafter refer to as \textit{asymptotic}, those which can only be guaranteed up to sampling uncertainty. 
\begin{theorem} \label{the:asymptotic_properties} Likelihood-based Bayesian regression markets of this kind yield the following asymptotic market properties:
\begin{enumerate}
    \item Individual rationality---support agents have a weak preference for participating in the market rather than not participating, that is, $\pi_{a, t} \geq 0, \forall a \in \mathcal{A}_{-c}, \ \forall t$.
    \item Zero-element---a support agent that provides no feature, or only provides features with zero marginal contribution to all coalitions of other features should receive no payment, that is, $\forall \mathcal{C} \in \mathcal{I}_{-c} : v_t (\mathcal{C}\cup \mathcal{I}_c \cup \{i\}) \equiv v_t (\mathcal{C} \cup \mathcal{I}_c), \ \forall i \in \mathcal{I}_{a} \mapsto \pi_{a} = 0, \ \forall t.$
    \item Truthfulness---support agents receive their maximum potential payment when reporting their true data, that is, $v_t (\mathcal{C} \cup \mathcal{I}_c ; \mathbf{x}_{\mathcal{C} \cup \mathcal{I}_c, t}) \geq v_t (\mathcal{C} \cup \mathcal{I}_c ; \mathbf{x}_{\mathcal{C} \cup \mathcal{I}_c} + \boldsymbol{\eta}_t), \ \forall \mathcal{C} \in \mathcal{I}_{-c}, \forall i \in \mathcal{C}_{-b}, \ \forall t$, where $\boldsymbol{\eta}_t$ represents noise added to the original feature.
\end{enumerate}
\noindent {\bf Proof} Individual rationality follows directly from Assumption~\ref{as:feature_selection} and zero-element follows directly from the null-player axiom of semivalues satisfied by the Shapley value. For a proof of truthfulness, see Appendix~\ref{app:truthfulness}.
\hfill\BlackBox
\end{theorem}

In practice, only an in-sample estimate of the posterior moments are available.
We assume that the specified hypothesis is such that as more data is observed, the posterior distribution converges to the Dirac measure around the maximum likelihood estimate of the parameter values almost surely, that is
\begin{align}
    D_{\textrm{KL}}( p(\Theta_{\mathcal{C}} \vert \mathcal{D}_{\mathcal{C}, t} \vert \vert \delta(\Theta^{\ast}_{\mathcal{C}}))
    \xrightarrow{\enskip t\enskip} 0, \quad \forall \mathcal{C},
    \label{eq:consistency}
\end{align}
where $\delta(\cdot)$ is the probability density function of the Dirac delta distribution and $\Theta^{\ast}$ is the maximum likelihood estimate of the parameters.
\begin{remark}
    This assumption implies asymptotic consistency of well-specified models. Although in practice model misspecification is inevitable, concentration around the maximum likelihood estimate is sufficient to guarantee the properties in Theorem~\ref{the:asymptotic_properties} hold up to sampling uncertainty.
\end{remark}

Given (\ref{eq:consistency}), individual rationality proceeds from Assumption~\ref{as:feature_selection}, as given $\phi_{i, t} \geq 0, \, \forall i \in \mathcal{I}_{-c}, \, \forall t$, it follows from definitions (\ref{eq:buyer_payment}) and (\ref{eq:seller_payment}) that payments can only be nonnegative in expectation. Similarly, the zero-element property, inherited from the null-player axiom, holds by design---if no feature is reported to the market then trivially no revenue is allocated, and if instead the true coefficient associated with a feature is zero, so too would be the associated revenue. 
Truthfulness ensures incentive compatibility, such that there is an incentive for support agents to report their true feature data. We assume that if a support agent is to provide an untruthful report of their data, they do so by the addition of centred noise with finite variance. Noise added to a particular feature is uncorrelated with noise added to any other, and conditionally independent of the target given the feature. 
\begin{corollary}
\label{cor:truthfulness} Following Assumptions~\ref{as:scoring_rule} and \ref{as:gaussian_hypothesis} the revenue of each of the support agents exhibits a unique maximum when each reports their true feature data.

\vspace{3mm}
\noindent {\bf Proof} See Appendix~\ref{app:truthfulness}.
\hfill \BlackBox
\end{corollary}

\begin{remark}
    Even in expectation, Theorem~\ref{the:asymptotic_properties} can only be guaranteed in-sample, and may not generalize to the out-of-sample market stage.
    This issue pertains to the rich field of generalization in machine learning, for which bounds can typically only be attained under strict assumptions about the data generating processes \citep{mohri2018foundations}. We leave a thorough examination of the generalization characteristics of these market properties to future work.
\end{remark}

Lastly, we acknowledge properties of similar markets proposed in related works. For instance, \citet{lambert2008self} introduce \textit{normality} in the context of wagering mechanisms, which would hold universally in our setup if features are independent. 
The same authors also introduce \textit{sybilproofness} and \textit{monotonicity}, which are deemed irrelevant to our setup. 
Another property frequently discussed in literature is that of \textit{robustness to replication}, which states that no support agent should be able to increase their revenue by replicating their data and submitting these replicates to the market under false identities. Whilst several mechanism designs have been proposed to satisfy this property (e.g., \citealt{agarwal2019marketplace}, \citealt{ohrimenko2019collaborative}, \citealt{han2022replication}), its satisfaction generally comes at a cost, for instance \citet{agarwal2019marketplace} sacrifice budget balance.
Therefore, data replication remains an open challenge; we leave exploration of this topic in relation to our setup as future work.

\subsection{Information-based Designs} \label{sec:information_designs}
Since the properties in Theorem~\ref{the:asymptotic_properties} can only be guaranteed in expectation, it is likely that they will be violated in a single shot of the market.
Whilst violating these properties would have no impact to the central agent with respect to predictive performance, support agents would be exposed to considerable financial risks, especially when a limited number of observations, as sub-optimal estimates of the parameters distorts allocations.
This issue would be exacerbated out-of-sample, for which the in-sample estimate of the posterior is likely to be less efficient.

To alleviate these risks, we explore alternative methods for valuing coalitions of features. Our approach is inspired by recent works concerned with multi-class classification with model outputs as discrete probability distributions. In this setting, \citet{covert2020understanding} demonstrate that models can be compared using relative mutual information. However this requires explicit computation of the joint distribution over the observed data, which may be intractable when dealing with continuous distributions, demanding expensive approximation \citep{kraskov2004estimating}.
Rather than focusing on predictive performance, in the work of \citet{agussurja2022convergence} multiple data owners seek to perform joint inference of a set of parameters. 
Each subset of features is valued using the information gain on the \textit{true} parameters, measured by the KL divergence of the posterior from a common prior. This is not immediately applicable to our setup, as we instead compensate support agents based on their contribution to overall predictive performance. 
Instead, we can make use of the information gain by considering the predictive densities, which encapsulate the value of the features in relation to predictive performance. In the following, we derive two methods for employing the KL divergence in our setup, demonstrating the implications on the market properties for each.

\subsubsection{Marginal Contribution}
We can express the marginal contribution of a feature to a coalition as the additional information that it provides, that is, the KL divergence between the predictive distribution \textit{with} and \textit{without} the particular feature, such that
\begin{align}
    m_{i, t}(\mathcal{C}) = \mathbb{E} [ D_{\textrm{KL}} (  p(y_{t}  \vert \textbf{x}_{\mathcal{C}  \cup \{i\}, t})  \vert \vert p(y_{t}  \vert \textbf{x}_{\mathcal{C} , t}) )], \quad \forall i, \ \forall \mathcal{C}.
    \label{eq:kld_marginal_contribution}
\end{align}
with $\mathcal{M}^{\mathrm{BLR}}_{\mathrm{KL}-m}$ denoting the corresponding market design.
We remove the conventional characteristic function altogether and replace it with a function that maps the predictive density of both coalitions to a real-valued scalar. With Assumption~\ref{as:gaussian_hypothesis}, we can express the KL divergence as the expected value of the logarithm of the Radon-Nikodym derivative, since any two univariate Gaussian distributions satisfy absolute continuity.
\begin{corollary}
\label{cor:kld_marginal_contribution_asymptotic} The definition in (\ref{eq:kld_marginal_contribution}) yields revenue allocations asymptotically equivalent to those obtained using the likelihood-based design.

\vspace{3mm}
\noindent {\bf Proof} See Appendix~\ref{app:kld_marginal_contribution_asymptotic}.
\hfill \BlackBox
\end{corollary}

Despite this asymptotic equivalence, the impact of using the KL divergence as described in (\ref{eq:kld_marginal_contribution}) becomes apparent when the number of observations is limited; the resultant revenue allocations will be less volatile, reducing risk exposure of the support agents. This results from the fact that the KL divergence accounts only for the relative entropy, considering the overall information held within the distributions rather than the specific observations of the target signal, which can be distorted by outliers. 
\begin{theorem} \label{the:kld_marginal_contribution} Replacing the marginal contribution with the definition in (\ref{eq:kld_marginal_contribution}) alters the market properties in Theorems~\ref{the:universal_properties} and \ref{the:asymptotic_properties} as follows: individual rationality becomes a universally held property at the expense of budget balance violation, whilst the remaining properties are unchanged. 

\vspace{3mm}
\noindent {\bf Proof} Individual rationality follows directly from Gibbs' inequality. For a proof of the loss of budget balance, see Appendix~\ref{app:budget_balance_violation}.
\hfill \BlackBox
\end{theorem}

Although using the KL divergence in such a way yields universal individual rationality by design,  reducing the definition of marginal contribution to a single inseparable expression removes the telescoping sum structure of the original Shapley value. This leads to a violation of the efficiency axiom and hence budget balance.
For brevity, we omit a proof for the remaining properties by virtue of similarity to Theorems~\ref{the:universal_properties} and \ref{the:asymptotic_properties}. Although budget balance is violated, the universal satisfaction of individual rationality theoretically removes the most severe financial risks exhibited by the support agents, as they are guaranteed a nonnegative revenue. We see this as a similar trade-off exhibited in \citet{agarwal2019marketplace} in pursuit of robustness to replication, that is, the addition of financial security is simply paid for by the market.

\subsubsection{Characteristic Function}
In some cases, violating budget balance may be impractical. If so, the KL divergence can instead be used in a manner that more closely resembles that presented in \citet{agussurja2022convergence}, however instead of considering the posterior distribution, we set the common prior to be the predictive distribution of the central agent, such that
\begin{align}
    v_t(\mathcal{C}) = \mathbb{E} [ D_{\textrm{KL}}(  p(y_{t} \vert  \textbf{x}_{\mathcal{C}, t})  \vert \vert p(y_{t}  \vert \textbf{x}_{\mathcal{I}_c , t}) ) ], \quad \forall i, \ \forall \mathcal{C}.
    \label{eq:kld_characteristic_function}
\end{align}

Now we have instead only modified the characteristic function, with $\mathcal{M}^{\mathrm{BLR}}_{\mathrm{KL}-v}$ denoting the corresponding market design. The marginal contribution is still $m_{i, t}(\cdot) = v_t (\cdot) - v_t (\cdot \cup \{i\})$. 
\begin{theorem} \label{the:kld_characteristic_function_asymptotic} Valuing a coalition as in (\ref{eq:kld_characteristic_function}) yields revenue allocations asymptotically equivalent to those obtained using the likelihood-based design.

\vspace{3mm}
\noindent {\bf Proof} See Appendix~\ref{app:kld_characteristic_function_asymptotic}
\hfill \BlackBox
\end{theorem}

\begin{corollary} \label{cor:kld_characteristic_function_asymptotic} Replacing the characteristic function of likelihood-based design with the definition (\ref{eq:kld_characteristic_function}) preserves the market properties in Theorems~\ref{the:universal_properties} and \ref{the:asymptotic_properties}. 

\vspace{3mm}
\noindent {\bf Proof} Omitted due to similarity to those for these theorems.
\hfill \BlackBox
\end{corollary}

Since the telescoping sum structure of the Shapley value remains, budget balance is reinstated as a universal property. However, individually rationality reduces back to an asymptotic property. This follows from the fact that the marginal contribution now involves subtraction of expected KL divergences, for which Gibbs’ inequality no longer applies. Nevertheless, this design should still provide us with less volatile allocations relative to the NLL-based designs when a limited number of observations are available. Hence, one should still expect reductions in risk exposure for the support agents, the extent to which will be studied through a series of simulation studies in Section~\ref{sec:simulation_studies}.

\subsection{Summary of Market Designs}
Apart from the transition to Bayesian regression analyses, the proposed market designs differ solely in their marginal contribution formulations. Both the payment of the central agent \textit{and} the revenue allocations are impacted by the difference in the functional form of $m_{i, t}$, the extent of which will also be studied in Section~\ref{sec:simulation_studies}. To end this section, we summarize the different formulations in Table~\ref{tab:summary_of_market_designs}.
\begin{table}[!ht]
    \centering
    \renewcommand{\arraystretch}{1.5}
    \begin{tabular}{ll}
        \toprule
        Market Design & Formulation of marginal contribution: $m_{i, t} (\mathcal{C}), \, \forall t, \ \forall i, \ \forall \mathcal{C}$ \\
        \midrule
        (i) $\mathcal{M}^{\mathrm{MLE}}_{\mathrm{NLL}}$ & $\mathbb{E} [-\log ( p(y_{t} \vert \textbf{x}_{\mathcal{C}, t}; \Theta^\ast_{\mathcal{C}} )) ] - \mathbb{E} [-\log ( p(y_{t} \vert \textbf{x}_{\mathcal{C} \cup \{i\}, t}; \Theta^\ast_{\mathcal{C} \cup \{i\}}) ) ]$ \\
        (ii) $\mathcal{M}^{\mathrm{BLR}}_{\mathrm{NLL}}$ & $\mathbb{E} [-\log ( p(y_{t} \vert \textbf{x}_{\mathcal{C}, t}) ) ] - \mathbb{E} [-\log ( p(y_{t} \vert \textbf{x}_{\mathcal{C} \cup \{i\}, t}) ) ]$ \\
        (iii) $\mathcal{M}^{\mathrm{BLR}}_{\mathrm{KL}-m}$ & $\mathbb{E} [ D_{\textrm{KL}} (  p(y_{t} \vert \textbf{x}_{\mathcal{C} \cup \{i\}, t}) \vert \vert p(y_{t} \vert \textbf{x}_{\mathcal{C} , t}) ) ]$ \\
        (iv) $\mathcal{M}^{\mathrm{BLR}}_{\mathrm{KL}-v}$ & $\mathbb{E} [ D_{\textrm{KL}} ( p(y_{t} \vert \textbf{x}_{\mathcal{C}, t}) \vert \vert p(y_{t} \vert \textbf{x}_{\mathcal{I}_c , t}) ) ] - \mathbb{E} [ D_{\textrm{KL}} ( p(y_{t} \vert \textbf{x}_{\mathcal{C} \cup \{i\}, t}) \vert \vert p(y_{t} \vert \textbf{x}_{\mathcal{I}_c , t}) ) ]$ \\
        \bottomrule
    \end{tabular}
    \caption{Marginal contribution formulation for each of the market designs introduced in Section~\ref{sec:market_properties}.}
    \label{tab:summary_of_market_designs}
\end{table}

\section{Simulation Studies} \label{sec:simulation_studies}
To illustrate our findings, we now present a collection of scenarios and simulation-based case studies.\footnote{Our code is publicly available at: \url{https://github.com/tdfalc/regression-markets}}
To emphasize the versatility of our proposed Bayesian regression market, we devote particular attention to four distinct setups, each portraying an additional layer of complexity to emulate real-world intricacies.
We note these setups provide simplified representations of the real world for the purpose of demonstration.
We explore compounding effects of likelihood misspecification, specifically with respect to both the interpolated function and the intrinsic noise in the target signal. 

In each of the simulation-based case studies, the central agent seeks to model a target variable $Y_t$ using their own feature $x_{1, t}$ and the relevant features available in the market, each owned by a unique support agent, namely $x_{2, t}$ and $x_{3, t}$. The likelihood is an independent Gaussian stochastic process with finite precision $\xi_{Y_t}$. The linear interpolant for the grand coalition is $f(\mathbf{x}_{t}, \mathbf{w}) = w_0 + w_1 x_{1, t} + w_2 x_{2, t} + w_3 x_{3, t}, \, \forall t$. The various setups differ solely in the model of the likelihood as follows: (i) \textit{Baseline}---the likelihood is well specified with respect to the \textit{true} data generating process, given by $p(y_t \vert \mathbf{x}_{t}, \mathbf{w}) = \mathcal{N}(f(\mathbf{x}_{t}, \mathbf{w}), \xi_{Y_t})$; (ii) \textit{Interpolant}---the interpolant is misspecified such that we write the \textit{true} mean of the likelihood as $f(\mathbf{x}_{t}, \mathbf{w}) = \textbf{w}^\top \left( \textbf{x}_t \odot \textbf{x}_t\right), \, \forall t$, where $\odot$ denotes the Hadamard product; (iii) \textit{Noise}---further to the misspecified interpolant, the Gaussian noise assumption is incorrect, with the \textit{true} process given by a Student’s t-distribution with two degrees of freedom; and (iv) \textit{Heteroskedasticity}---the non-Gaussian noise is heteroskedastic, such that at each time step it is multiplied by $x^2_{2, t}$.

\subsection{In-sample Market}
\begin{figure}[t]
\centerline{\includegraphics[width=1\textwidth]{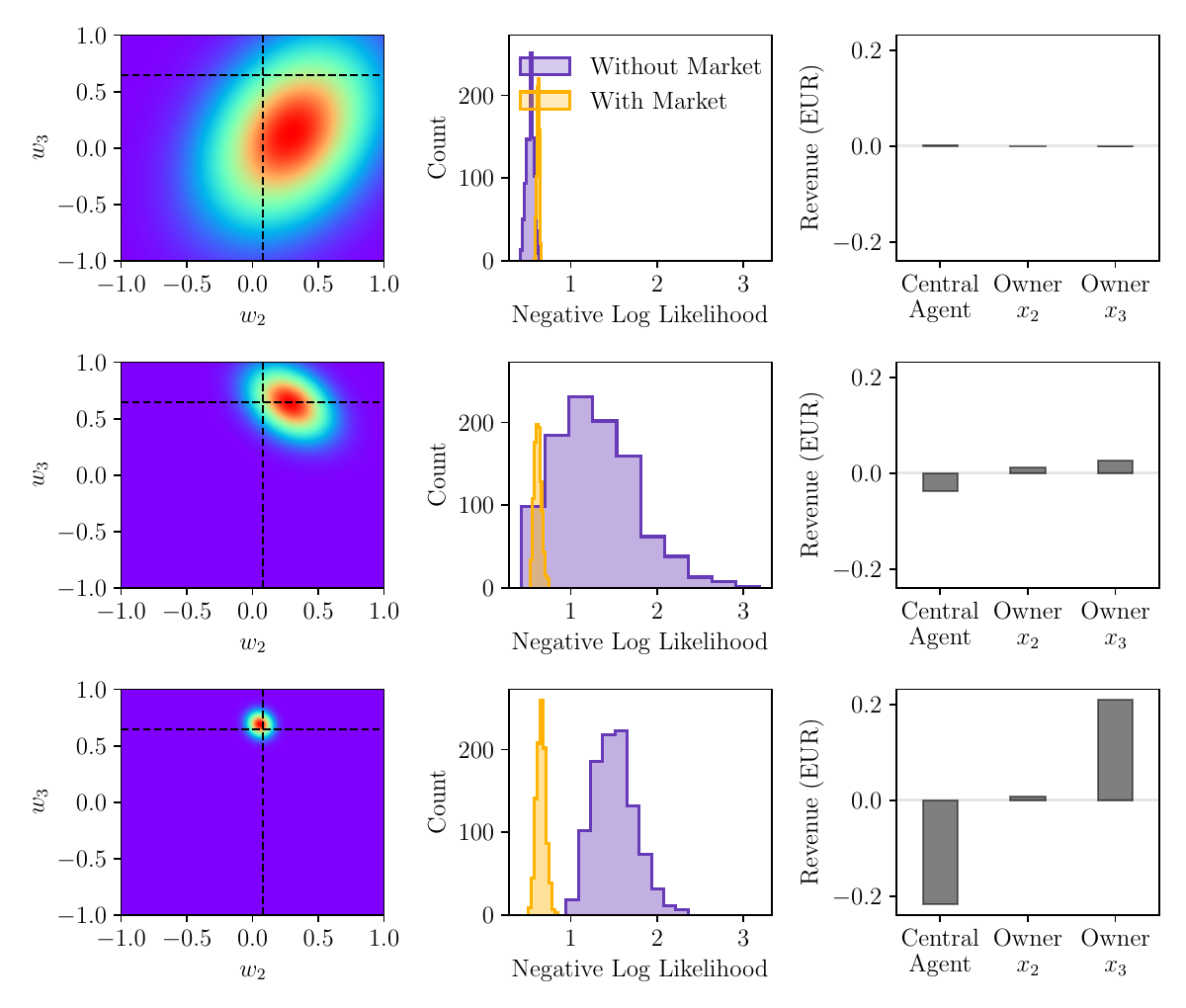}}
    \hspace{12mm}
    \begin{subfigure}[]{0.25\textwidth}
        \centering
        \caption{Posterior Updates}
        \label{fig:market_outcomes_a}
    \end{subfigure}
    \hspace{7mm}
    \begin{subfigure}[]{0.3\textwidth}
        \centering
        \caption{Predictive Performance}
        \label{fig:market_outcomes_b}
    \end{subfigure}
    \hspace{5mm}
    \begin{subfigure}[]{0.25\textwidth}
        \centering
        \caption{Revenue}
        \label{fig:market_outcomes_c}
    \end{subfigure}
     \caption{In-sample market with increasing batch size. The dashed lines in (a) highlight the \textit{true} coefficients. The histogram in (b) shows the in-sample NLL distribution. The bars in (c) are the cumulative revenues given the value of each datapoint provided.}
    \label{fig:market_ouctomes}
\end{figure}

We first demonstrate the link between Bayesian inference with increasing sample size and the eventual revenue allocation, using the in-sample stage of $\mathcal{M}_{\textrm{MLE}}^{\textrm{BLR}}$ as case study. We emulate batch inference (i.e., $\tau = 1$) for simplicity and consider only the \textit{Baseline} setup.
We assume the \textit{true} coefficients to be $\textbf{w} = [-0.11, 0.31, 0.08, 0.65]^\top$, and the noise precision to be constant for all time steps, treated as a hyperparameter with $\xi_{Y_t} = 3.31, \, \forall t$. We further set the valuation of the central agent to $\lambda = 0.01$ EUR per time step and per unit improvement in $\ell$. We run the market for increasing sample sizes, specifically $5$, $10$ and $50$, recording posterior moments, predictive performance and market revenue allocations for each. The results are shown in Figure~\ref{fig:market_ouctomes}.

In Figure~\ref{fig:market_outcomes_a}, we see that as one would expect, increasing the number of observations improves the estimation of the posterior, eventually centering around the \textit{true} coefficient values. In Figure~\ref{fig:market_outcomes_b}, we present the distribution of $10^3$ sub-sampled NLLs. In the first case, the predictive performance seemingly gets worse using the regression market, rendering small and possibly negative revenues for the support agents. However, this is simply due to overfitting and emphasizes the importance of Assumption~\ref{as:feature_selection}, as a prior feature selection process could remove these features to preserve individual rationality. As the sample size increases, the improved posterior facilitates better capturing of the information provided by the features of the support agents, for which the central agent must pay for, highlighted by the extra revenue earned by the support agents, shown in Figure~\ref{fig:market_outcomes_c}.

\subsection{Uncertainty Quantification} \label{sec:uncertainty_quantification}
Next we illustrate our four considered setups, highlighting the benefit to the central agent of merely facilitating Bayesian regression analyses. 
We set the \textit{true} parameters to $\textbf{w} = [-0.1, 0.3, 0.8, -0.4]^\top$ and $\xi_{Y_t} = 0.5, \, \forall t$. We again emulate batch inference and run a Monte-Carlo simulation whereby we clear the market $10^3$ times for several different sample sizes and record the expected NLL for $10^3$ out-of-sample observations for each.
This is carried out using both maximum likelihood estimation and Bayesian regression analyses.

\begin{figure}[t]
\centerline{\includegraphics[width=1\textwidth]{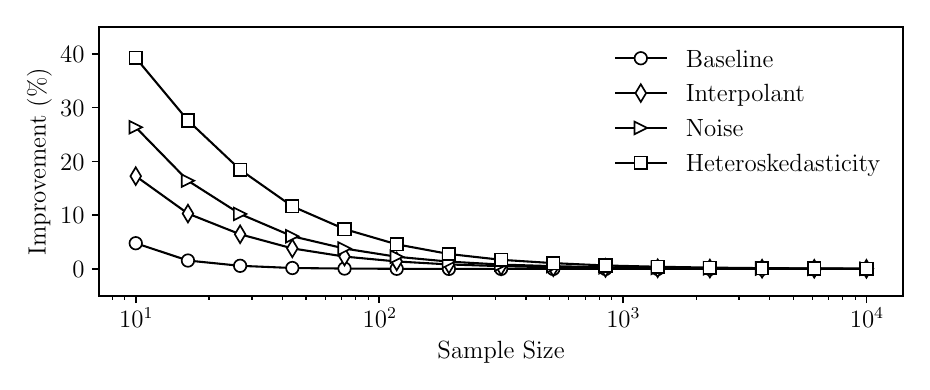}}
    \caption{Empirical average of the percentage improvement in the NLL ratio for BLR relative to MLE, plotted as a function of sample size.}
    \label{fig:uncertainty_quantification}
\end{figure}

Figure~\ref{fig:uncertainty_quantification} shows the empirical average of the percentage improvement in the objective value for the Bayesian regression model. Observe that the improvement is most significant across all setups when the sample size is relatively small, as the additional piece of uncertainty in the parameter estimates plays a greater role in the predictive distribution, increasing the predictive likelihood. Then, as the sample size increases, the parameter estimates converge in accordance with (\ref{eq:consistency}). Furthermore, as the additional layers of complexity are introduced, the benefit of incorporating parameter uncertainty increases considerably. These improvements attained by converting to a Bayesian framework indicate a better calibration of uncertainty, enriching the information used by the central agent for risk-informed decision-making downstream, compared with the deterministic proposals of \citet{agarwal2019marketplace} and \citet{pinson2022regression}.

\subsection{Convergence Analysis} \label{sec:convergence_analysis}
\begin{figure}[t]
\centerline{\includegraphics[width=1\textwidth]{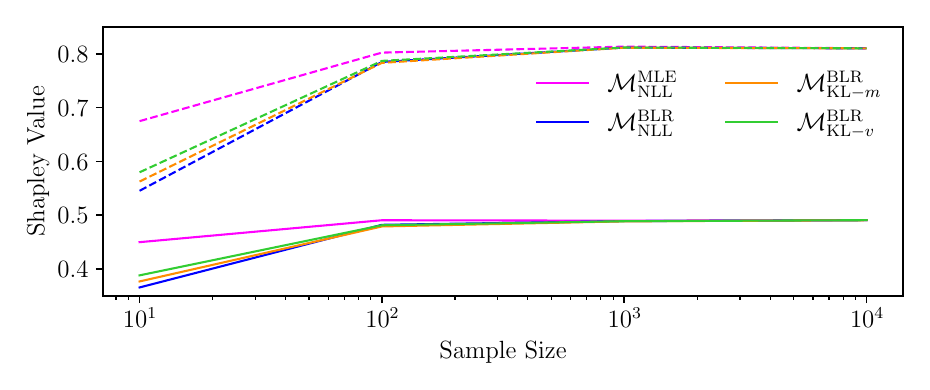}}
    \caption{Empirical average of expected Shapley values for each market design, plotted as a function of sample size. Solid and dashed lines correspond to features $x_{2, t}$ and $x_{3, t}$, respectively.}
    \label{fig:shapley_convergence}
\end{figure}

\begin{figure}[t]
\centerline{\includegraphics[width=1\textwidth]{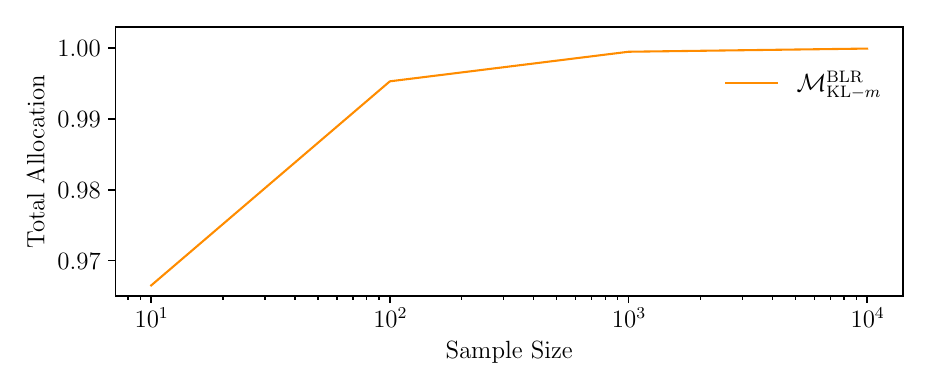}}
    \caption{Empirical average of the expected total allocation for the $\mathcal{M}_{\mathrm{KL}-m}^{\mathrm{BLR}}$ market design. Only this design is plotted since, in the remaining markets, budget balance is a universal property by design hence the total allocation is always 1.}
    \label{fig:allocation_convergence}
\end{figure}

Now we present an empirical study of the in-sample asymptotic convergence for our various market designs. 
Let the \textit{true} coefficeints and noise precision be given by $\textbf{w} = [-0.1, 0.8, 0.7, -0.9]^\top$ and $\xi_{Y_t} = 1.0, \, \forall t$, respectively, focusing here solely on the \textit{Baseline} setup, since asymptotic convergence is irrespective of the \textit{true} data generating processes, but rather the set of modelling assumptions. A similar Monte-Carlo simulation is performed, recording the in-sample Shapley values for each run, the results of which are presented in Figure~\ref{fig:shapley_convergence} and Figure~\ref{fig:allocation_convergence}.

Looking first at Figure~\ref{fig:shapley_convergence}, we see that with a small sample size, the frequentist market design assigns a larger contribution to the features compared with those using Bayesian regression, however these values indeed converge asymptotically in align with the theory. This discrepancy is likely due to the greater reduction in the in-sample objective provided by the maximum likelihood estimate, which is of course prone to overfitting. 
In Figure~\ref{fig:allocation_convergence}, we observe that the $\mathcal{M}^{\mathrm{BLR}}_{\mathrm{KL}-m}$ market renders a surplus in revenue when the sample size is small.
This demonstrates the trade-off incurred by virtue of the now universally held individual rationality property---budget balance is no longer guaranteed, even during the in-sample stage. This problem does resolve with increasing number of observations as the Shapley values converge.

\subsection{Risk Exposure} \label{sec:risk_exposure}
We now turn our attention to the finances of the support agents, which we assess by computing both the expected value of the revenue, $\int \pi_{a, t} p(\pi_{a, t}) \, d\pi_{a, t}, \, \forall t$, and the expected shortfall (i.e., conditional value at risk), $-1/\alpha \int_{\pi_{a, t} \leq q_{\alpha}(\pi_{a, t})} \pi_{a, t} p(\pi_{a, t}) \, d\pi_{a, t}, \, \forall t$, for all $a \in \mathcal{A}_{-c}$, where $q_{\alpha}(\cdot)$ is the quantile with nominal level $\alpha \in (0, 1)$. We present empirical estimations of these financial metrics for a case study where we again clear the market for a new sample of data $10^3$ times and record the revenue of each support agent, with the \textit{true} coefficients set to $\textbf{w} = [0.1, -0.5, 0.0, 0.7]^\top$, with noise precision $\xi_{Y_t} = 0.67, \, \forall t$. We additionally set $\lambda = 0.03$ EUR per time step and per unit improvement in $\ell$ for the both in-sample and out-of-sample stages. We use a simple sub-sampling method to derive the corresponding two-sided confidence intervals of both the expected value and expected shortfall of the revenue with a 95\% confidence level. We run this simulation for each market design, as well as each misspecification setup, with $10^3$ in-sample and out-of-sample observations.

\begin{figure}[t]
\centerline{\includegraphics[width=1\textwidth]{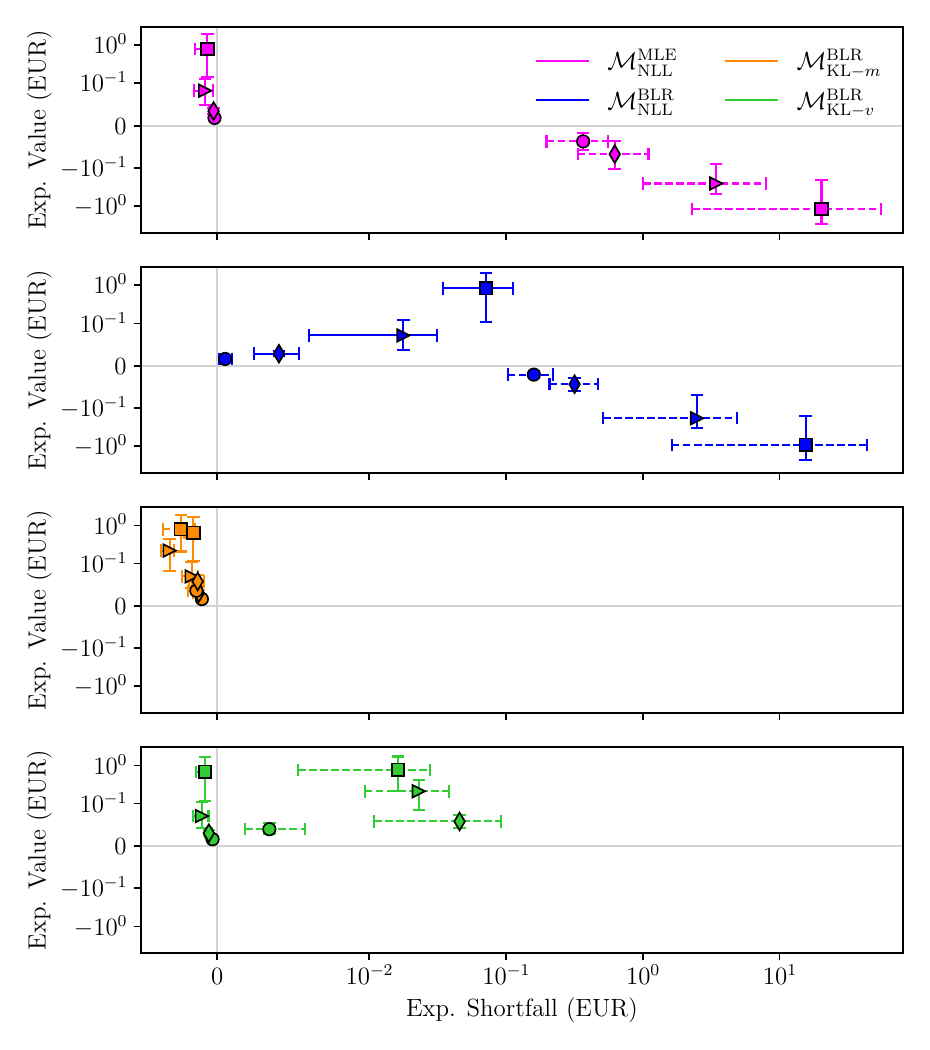}}
    \caption{Two-sided confidence intervals with a 95\% confidence level for both the expected value and expected shortfall of the revenue received by the owner of $x_{2, t}$, with quantile parameter $\alpha = 0.05$, for each setup, namely \textit{Baseline} ($\circ$), \textit{Interpolant} ($\diamond$), \textit{Noise} ($\triangleright$) and \textit{Heteroskedasticity} ($\square$). Both in-sample (solid) and out-of-sample (dashed) metrics are plotted.}
    \label{fig:revenue}
\end{figure}

In Figure~\ref{fig:revenue}, we plot the revenue of the support agent who owns $x_{2, t}$. Observe that the expected value of the revenue is relatively consistent across all market designs for each setup. Yet, for each additional layer of complexity, the expected shortfall is positive for the $\mathcal{M}^{\mathrm{BLR}}_{\mathrm{NLL}}$ market, increasing by almost two orders of magnitude in the latter setups. For the designs that use the KL divergence, the expected shortfall remains somewhat constant around zero, highlighting the sizeable reductions in risk exposure compared to using NLL. As before, the $\mathcal{M}^{\mathrm{MLE}}_{\mathrm{NLL}}$ market design seemingly performs better than its Bayesian counterpart in-sample, however this does not generalize out-of-sample.
For this small sample size, parameter estimates are more likely to be sub-optimal, and hence the predictive likelihood is more volatile. In consequence, even the expected value of the out-of-sample revenue becomes increasingly negative with additional layers of complexity for the likelihood-based market designs, meaning that the individual rationality property is violated even in expectation. The out-of-sample revenues in the $\mathcal{M}^{\mathrm{MLE}}_{\mathrm{NLL}}$ market become worse than for $\mathcal{M}^{\mathrm{BLR}}_{\mathrm{NLL}}$, demonstrating that the superior performance of this market in-sample was indeed due to overfitting. In contrast, the expected value of the revenue is relatively consistent with those in-sample for both KL divergence-based markets. 

The expected shortfall for the likelihood-based markets increases by several orders of magnitude in the out-of-sample stage. Interestingly, one can now observe the consequence of re-instating budget balance by modifying the characteristic function instead of the marginal contribution. Specifically, whilst for the $\mathcal{M}^{\mathrm{BLR}}_{\mathrm{KL}-m}$ market design individual rationality is held universally, the expected shortfall in $\mathcal{M}^{\mathrm{BLR}}_{\mathrm{KL}-v}$ drifts positive. That being said, the risks are generally much less compared with $\mathcal{M}^{\mathrm{BLR}}_{\mathrm{NLL}}$, implying there is still merit to this approach. These results demonstrate that valuing features based on the the information provided is able to mitigate risks entirely,
despite being asymptotically equivalent to proposals of \citet{agarwal2019marketplace} and \citet{pinson2022regression}.

\subsection{Nonstationary Processes} \label{sec:nonstationary_processes}
So far we have assumed only batch inference (i.e., $\tau = 1$), however in Section~\ref{sec:market_clearing} we showed that theoretically this is merely a specification of the more general online Bayesian inference problem, which facilitates time-varying posterior moments. For the final simulation-based case study, let us consider a nonstationary data generating process, wherein the parameters initially take on the values $\textbf{w} = [0.0, -0.2, 0.1, 0.3]^\top$, with noise precision $\xi_{Y_t} = 0.98, \, \forall t$. 

\begin{figure}[t]
\centerline{\includegraphics[width=1\textwidth]{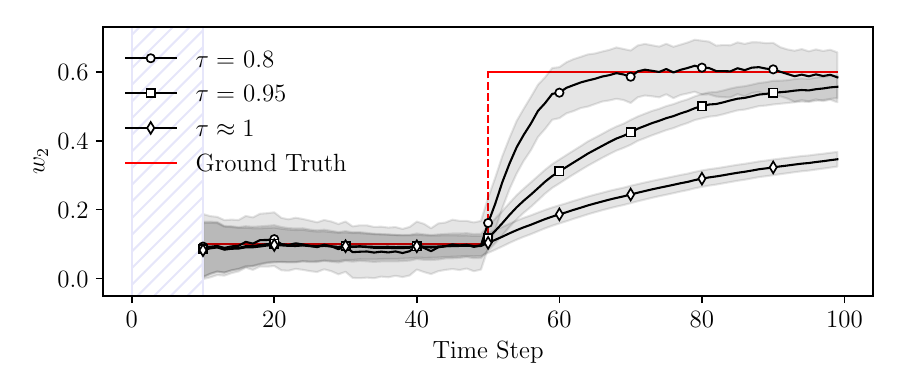}}
    \caption{Temporal evolution of the empirical average of the estimated value for $w_2$, with standard errors shaded gray. The estimates for the remaining parameters are omitted for clarity. The hatched area indicates the burn in period.}
    \label{fig:nonstationary_processes}
\end{figure}

For simplicity, we only let the coefficient associated with $x_{2, t}$ vary with time, with the remaining kept constant. To illustrate the effect of likelihood flattening, we consider cases where $w_2$ exhibits a discontinuity, representing a much more complex processes to capture with respect to its stationary analogue. We carry out a Monte-Carlo simulation whereby we record the empirical average of the parameter estimates at each time step with various values for $\tau$, the results of which are presented in Figure~\ref{fig:nonstationary_processes}. Of course, for the previous time-invariant cases, there would be no advantage of using likelihood flattening since the coefficients are stationary. For the more complex cases, as $\tau \mapsto 1$, our posterior beliefs decay more gradually, but as $\tau$ is reduced, we are able to better track the coefficient values, albeit with increased variance due to the fact that more weight is given to the flat prior.

\begin{figure}[t]
\centerline{
    \includegraphics[width=1\textwidth]{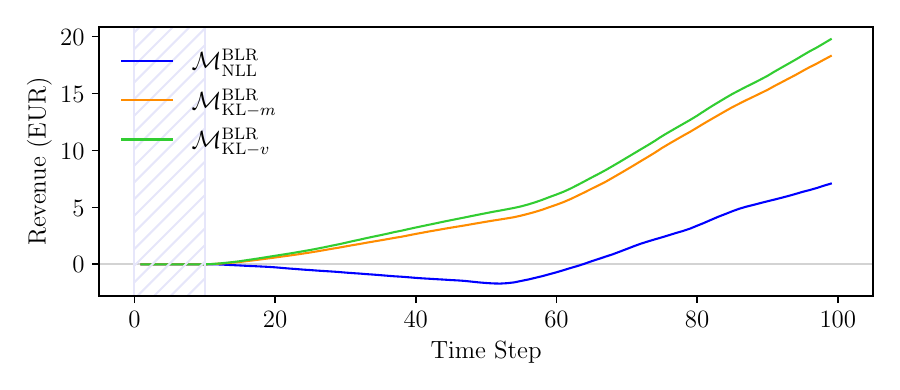}}    
    \caption{Cumulative empirical averages of the expected value of the revenue for the first support agent considering the \textit{Baseline} setup. The hatched area indicates the burn in period.}
    \label{fig:contributions_risk}
\end{figure}

We run a Monte-Carlo simulation whereby we fix $\tau = 0.95$, a trade-off between better tracking of the coefficient and increased variance. We re-run the market clearing procedure $10^3$ times, each time tracking the temporal evolution of the out-of-sample market revenue over $10^2$ time steps. We set $\lambda = 0.95$ EUR per time step and per unit improvement in $\ell$ and carry out this simulation for each of the proposed Bayesian regression market designs, considering only the \textit{Baseline} setup, the results of which are shown in Figure~\ref{fig:contributions_risk}. 

Given the use of likelihood flattening, each of the market designs are able to capture the step-change in the \textit{true} coefficient value. However, the extent of flattening required reduces the effective window size of observations, emulating a consistently small sample size even as more observations arrive. As a result, the likelhood-based market exhibits poor generalization. In fact, even though the true coefficient is $w_2 = 0.1$ before the step change, the expected value of the revenue is less than $0$, resulting in a negative cumulative revenue in the first half of the simulation, and hence the overall revenue earned by the agent is considerably less. This highlights that valuing features using information gain has the propensity to mitigate financial risks entirely, even out-of-sample where the majority of risk exposure manifests, compared with likelihood-based market designs in \citet{agarwal2019marketplace} and \citet{pinson2022regression} which subject agents to sizeable losses.

Note that here we have investigated concept drift with respect to the model parameters. Specifically, we show that our fully online setup is as robust as possible to such drift with the parameter estimates being updated at the closest time step to the next, rather than training the model once and making predictions for several time steps (e.g., in batch or rolling window inference). This indeed has a greater computational cost, but exploring this trade-off is outwith the scope of this work. In addition, what we have not explored is concept drift with respect to the model hyperparameters (e.g., the basis function vector, the noise process, etc.). This of course is a key problem in real forecasting applications, however in our setup it is the central agent who has discretion over these hyperparameters. An interesting direction for future work would be to entrust the market operator also with such model selection, in which case it would be up to the market to account for hyperparameter drift as well.

\section{Real-world Application} \label{sec:real_world}
We complete our experimental analysis by verifying the applicability of our proposal to real-world applications. We make use of an open source dataset, namely the \textit{Pan-European Climate Database}, as detailed in \cite{koivisto2022data}. This dataset consists of hourly average irradiance values for European countries, obtained by simulating the output from south-facing solar photovoltaic (PV) modules across several intra-country regions. Although this data is not exactly \textit{real}, it effectively captures the spatio-temporal aspects of solar irradiance across the continent, with the benefit of not being contaminated with spurious data points, as can often be the case with real-world datasets. 

Suppose that the electricity system operator in each country seeks to forecast its own country's average generation from solar PV modules, with a view to estimate electricity demand and determine balancing resource requirements.
For illustration, we consider six countries: United Kingdom (UK), Belgium (FR), Austria (AT), Greece (GR), Cyprus (CY) and Turkey (TR), each of which is assumed to enter the regression market to enhance their respective forecasts.
For simplicity, we focus on a 1-hour forecasting horizon (i.e., nowcasting) using only linear basis functions, though both longer latency periods and more complex models could be considered. 

\begin{figure}[t]
\centerline{\includegraphics[width=1\textwidth]{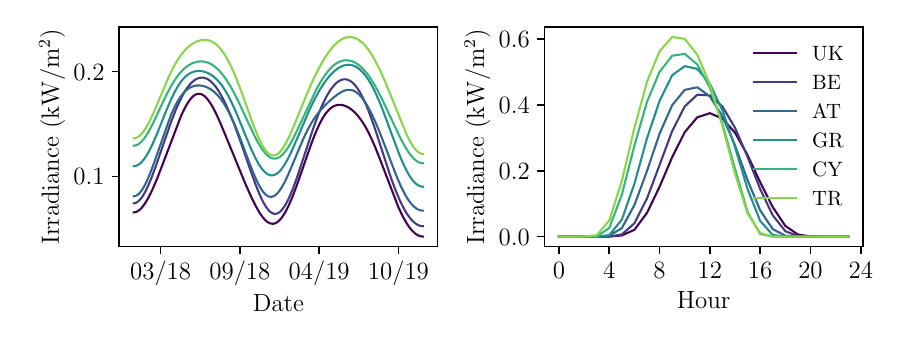}}
    \hspace{10mm}
    \begin{subfigure}[]{0.4\textwidth}
        \centering
        \caption{Rolling Average}
        \label{fig:rolling_avg}
    \end{subfigure}
    \hspace{14mm}
    \begin{subfigure}[]{0.4\textwidth}
        \centering
        \caption{Hourly Average}
        \label{fig:hourly_avg}
    \end{subfigure}
    \caption{The rolling average and hourly average solar irradiance observed in each country during the two-year time period of 2018--2019.}
    \label{fig:irradiance}
\end{figure}

We extract data that spans a two-year period from the start of 2018 to the end of 2019, with an hourly resolution. Suppose that each of the six countries takes turn in assuming the role of the central agent in parallel transactions. We use a simple \textit{Auto-Regressive with eXogenous input} model with a maximum of one lag for each feature. For solar energy, forecasting with lags simultaneously captures temporal correlations at particular locations and any indirect spatial correlations between neighboring locations, resulting from the natural development of cloud coverage and the rotation of the sun. We present the rolling average of the raw irradiance values in Figure~\ref{fig:rolling_avg}, which highlights the seasonality of generation, peaking during the summer months as expected. Similarly, by plotting the hourly average irradiance in Figure~\ref{fig:hourly_avg}, one can observe the spatial correlations such that at any given time, the actual generation in the more Easterly countries could be indicative of what is to come in Western Europe later in the day.

For each forecast, we model the likelihood as an independent Gaussian stochastic process with finite precision, similar to the setup in Section~\ref{sec:simulation_studies}. We consider an online setting such that over the entire two-year period, at each time step (i.e., one hour interval), when a new observation of the target signal is collected, the forecast issued at the previous time step is used for out-of-sample market clearing, whilst at the same time, the posterior is updated and the in-sample market is cleared, and a forecast for the next time step is subsequently made. We set $\tau = 0.998$ and assume the valuation of each central agent to be $\lambda = 50$ EUR and $\lambda = 150$ EUR per time step and per unit improvement in $\ell$ for the in-sample and out-of-sample stages, respectively, to reflect costs of balancing. With each country set as the central agent, we record the predictive performance and cumulative revenues of the remaining countries across both stages over the entire two-year period.

Let us first consider the improvements in predictive performance exhibited by each of the countries when assuming the role of the central agent. We present the average quarterly results in Table~\ref{tab:simulation_real}, using $\mathcal{M}^{\mathrm{BLR}}_{\mathrm{KL}-v}$ as a case study. In general, we observe a seasonality in the loss equivalent to that of the irradiance itself, such that smaller enhancements in predictive performance are exhibited during the end two quarters, since there is less potential for improving predictive performance when irradiance is low. Both United Kingdom and Greece receive the greatest improvements, with Cyprus and Turkey the smallest, the latter of which is likely due to the fact that these countries are further East, thus less able to exploit the spatial correlations depicted in Figure~\ref{fig:hourly_avg}. We also note that the distribution of performance improvements amongst the countries is fairly similar between the in-sample and out-of-sample stages, which suggests any nonstationarities, as well as the time-varying objective estimates, are smooth, and hence the in-sample posterior is a relatively efficient estimator for use out-of-sample in the next time step.

\begin{table}[!ht]
    \centering
    \begin{tabular}{lcccccccc}
        \toprule
        \multirow{2}{*}{Country} & \multicolumn{4}{c}{In-sample} & \multicolumn{4}{c}{Out-of-sample} \\
        & Q1 &  Q2 &  Q3 &  Q4 & Q1 & Q2 & Q3 &  Q4  \\
        \midrule
        UK & $0.40$ & $2.24$ & $2.24$ & $0.37$ & $0.39$ & $2.32$ & $2.45$ & $0.36$ \\
        BE & $0.34$ & $1.58$ & $1.59$ & $0.51$ & $0.33$ & $1.60$ & $1.61$ & $0.50$ \\
        AT & $0.66$ & $1.77$ & $1.47$ & $0.72$ & $0.65$ & $1.81$ & $1.49$ & $0.72$ \\
        GR & $0.73$ & $2.11$ & $2.40$ & $0.82$ & $0.74$ & $2.15$ & $2.44$ & $0.81$ \\
        CY & $0.44$ & $1.05$ & $1.20$ & $0.56$ & $0.43$ & $1.05$ & $1.21$ & $0.55$ \\
        TR & $0.43$ & $1.00$ & $1.35$ & $0.65$ & $0.42$ & $1.00$ & $1.36$ & $0.64$ \\
        \bottomrule
    \end{tabular}
    \caption{Fractional improvement in the NLL ratio by virute of participating in the regression market, averaged over each calendar quarter, for both in-sample and out-of-sample market stages.}
    \label{tab:simulation_real}
\end{table}

\begin{table}[!ht]
    \centering
    \begin{tabular}{lcccc}
        \toprule
        Country &
        $\mathcal{M}^{\mathrm{MLE}}_{\mathrm{NLL}}$ &  $\mathcal{M}^{\mathrm{BLR}}_{\mathrm{NLL}}$ & $\mathcal{M}^{\mathrm{BLR}}_{\mathrm{KL}-m}$ &  $\mathcal{M}^{\mathrm{BLR}}_{\mathrm{KL}-v}$  \\
        \midrule
        UK & $2.23$ & $2.23$ & $2.24$ & $2.23$ \\
        BE & $2.01$ & $2.01$ & $2.01$ & $2.01$ \\
        AT & $2.31$ & $2.31$ & $2.34$ & $2.33$ \\
        GR & $2.60$ & $2.60$ & $2.64$ & $2.63$ \\
        CY & $1.83$ & $1.83$ & $1.88$ & $1.85$ \\
        TR & $1.79$ & $1.79$ & $1.84$ & $1.83$ \\
        \bottomrule
    \end{tabular}
    \caption{Total out-of-sample payments ($\times 10^6$ EUR) made to the regression market by each country when acting as the central agent.}
    \label{tab:out_of_sample_revenue}
\end{table}

\begin{figure}[!t]
\centerline{\includegraphics[width=1\textwidth]{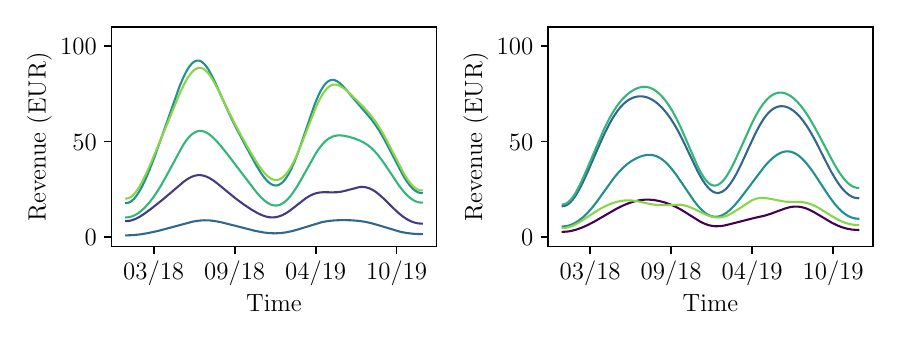}}
    \hspace{10mm}
    \begin{subfigure}[]{0.4\textwidth}
        \vspace{-10mm}
        \centering
        \caption{UK}
        \label{fig:uk}
    \end{subfigure}
    \hspace{14mm}
    \begin{subfigure}[]{0.4\textwidth}
        \vspace{-10mm}
        \centering
        \caption{BE}
        \label{fig:be}
    \end{subfigure}
\vskip -0.16 in
\centerline{\includegraphics[width=1\textwidth]{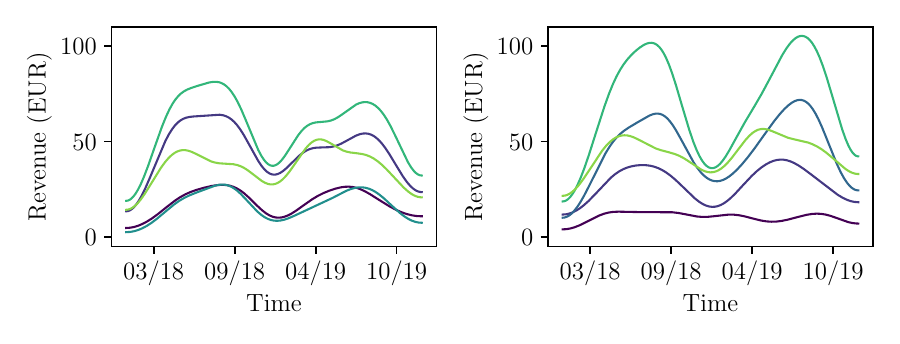}}
    \hspace{10mm}
    \begin{subfigure}[]{0.4\textwidth}
        \vspace{-10mm}
        \centering
        \caption{AT}
        \label{fig:at}
    \end{subfigure}
    \hspace{14mm}
    \begin{subfigure}[]{0.4\textwidth}
        \vspace{-10mm}
        \centering
        \caption{GR}
        \label{fig:gr}
    \end{subfigure}
\vskip -0.16 in
\centerline{\includegraphics[width=1\textwidth]{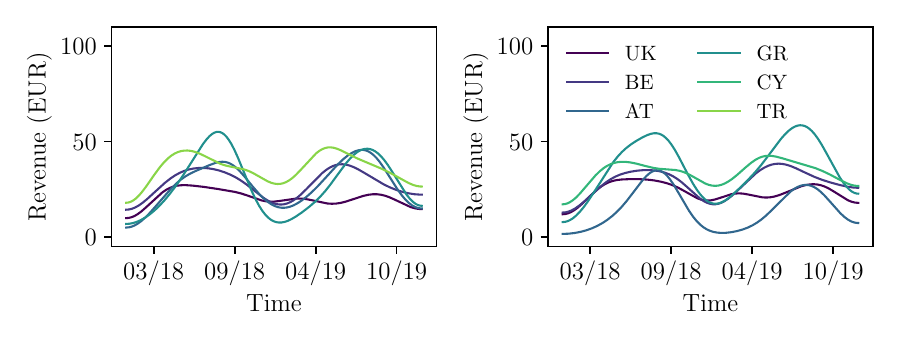}}
    \hspace{10mm}
    \begin{subfigure}[]{0.4\textwidth}
        \vspace{-10mm}
        \centering
        \caption{CY}
        \label{fig:cy}
    \end{subfigure}
    \hspace{14mm}
    \begin{subfigure}[]{0.4\textwidth}
        \vspace{-10mm}
        \centering
        \caption{TR}
        \label{fig:tr}
    \end{subfigure}
     \caption{Smoothed evolution of total revenue per time step earned by each of the six countries to the remaining five whilst assuming the role of the central agent.}
    \label{fig:simulation_real}
\end{figure}

In Figure~\ref{fig:simulation_real}, we present the smoothed evolution of the revenues across both the in-sample and out-of-sample market stages. As with the loss estimates, the allocation is by no means constant with time, such that the revenues of each agent are typically lower over the winter months and increase throughout the rest of the year. The value of each observation therefore also reflects the seasonality observed in the generation from solar PV modules. 
We also observe the spatio-temporal dynamics of solar irradiance, as countries to the East of the central agent, particularly those nearby or with high nominal generation, contribute most to the uplift. The revenues received by the remaining countries when either Cyprus or Turkey assume the role of the central agent are relatively small, in accordance with the results in Table~\ref{tab:simulation_real}. 

Lastly, we note that the revenues earned by some countries over the entire two-year period are substantial, for instance with Greece as the central agent, the system operator in Cyprus earns approximately $1.2 \times 10^6$ EUR, representing an average unit value of around $70$ EUR per observation shared. In Table~\ref{tab:out_of_sample_revenue}, we present the total out-of-sample payments collected from each country when acting as the central agent, for each market design. It is evident that the enhancements in predictive performance as a result of participating in the market has led to sizeable increases in revenue in the downstream decision-making process, with Easterly countries benefiting the most. As the number of in-sample observations is large, both $\mathcal{M}^{\mathrm{BLR}}_{\mathrm{KL}-m}$ and $\mathcal{M}^{\mathrm{BLR}}_{\mathrm{KL}-v}$ produce small differences in payments compared with the likelihood-based designs, that is, those similar to \citet{agarwal2019marketplace} and \citet{pinson2022regression}, which was expected given the asymptotic convergence of these market designs demonstrated in Figure~\ref{fig:shapley_convergence}. This is also the reason why $\mathcal{M}^{\mathrm{MLE}}_{\mathrm{NLL}}$ and $\mathcal{M}^{\mathrm{BLR}}_{\mathrm{NLL}}$ produce equivalent payments, described by the consistency result in (\ref{eq:consistency}).

\section{Conclusions} \label{sec:conclusion}
In modern industry, firms that employ predictive analytics (e.g., machine learning) often lack access to adequate datasets. Whilst sharing data amongst others could bring potential advantages, many firms remain hesitant to do so, mainly due to privacy concerns and the fear of losing a competitive edge, rather than practical complexities involved in establishing data sharing pipelines. Analytics markets, or in our case \textit{regression markets}, offer a possible solution to this, wherein data is commoditized with respect to the particular analytics task at hand, providing incentives for information exchange through remuneration. 

In this paper, we proposed a mechanism design for a regression market that facilitates a generalized approach to forecasting, one based on Bayesian regression analyses. As a result, we provide the buyer with richer and more nuanced information about future outcomes, offering better calibration of uncertainty to be used for risk-informed decision-making downstream. We first introduced what we posed as the Bayesian analogue of recent frequentist-based proposals, but showed that this market design, akin to those in current literature, exposes the buyer to considerable financial risks, especially when a limited number of observations are available or when the data generating processes are nonstationary. In these settings, sub-optimal estimates of the posterior distribution led to sizeable expected losses, especially during the out-of-sample market stage, for which the in-sample estimates of the posterior moments are less efficient.

To mitigate these risks, we formulated the value of a feature in terms of the information gain it provides. In particular, we derived two alternative definitions of the marginal contribution of a feature towards a set of other features using the KL divergence, the first of which could guarentee individual rationality universally (i.e., no support agents would be allocated negative revenue). There is of course no free lunch, as this was at the expense of budget balance. Nevertheless, we showed that in both cases that using the KL divergence was able to provide more robust revenue allocations by alleviating the financial risks, even at the out-of-sample market stage.

Possible directions for immediate future work could include extending the concepts of our proposal to a broader class of machine learning models, such as (i) non-convex regression, which will have implications on market property guarantees; (ii) non-Gaussian hypotheses, which may require approximation bounds; and lastly (iii) alternative modelling paradigms, for instance, classification, unsupervised learning, or data-driven optimization problems in general. 

On a broader note, there are still many unanswered questions in relation to the complexities of treating data as a commodity. For instance, in practice, datasets cover different spatial and temporal horizons, and may become (un-)available to the market at different times. Accordingly, aggregating real-world datasets in an online fashion may not be straightforward and may require revision of fundamental concepts in online learning and mechanism design. Additionally, much of the current literature relies on the assumption that the valuation of the central agent is both linear and easily conceivable with respect to the loss function, which may not be true if the downstream decision-making process is complex or in the face of externalities, for instance, whether or not competing firms also get access to the data may affect the valuation. Support agents may also have reservations to share their information, for instance due to privacy concerns or conflicts of interest. As well as physical costs of collecting and storing data, this may require a minimum revenue threshold to be established. Lastly, if firms that share data are indeed competitors in a downstream market, one may be interested in if, by faciliating better use of information, the analytics market is beneficial to overall social welfare, and if those that lose competitive advantage by sharing their information are adequately compensated.

\appendix
\section{Truthfulness Property (Theorem~\ref{the:asymptotic_properties})} \label{app:truthfulness}
We provide a proof of the truthfulness property provided by Theorem~\ref{the:asymptotic_properties}, which describes the asymptotic market properties.

We model the target signal, $\{Y_t\}$, as a deviation from the deterministic linear interpolant in (\ref{eq:interpolated_function}) under the following centred additive noise process:
\begin{equation*}
    p( \mathcal{D}_{\mathcal{C}, t} \vert \Theta_{\mathcal{C}}) = \prod_{t^{\prime} \leq t} \mathcal{N} (f(\textbf{x}_{\mathcal{C}, t}, \textbf{w}_{\mathcal{C}}), \xi_{Y_t}^{-1}), \quad \forall t. 
\end{equation*}
Without loss of generality, consider $\xi_{Y_t} = \xi, \forall t$  to be a hyperparameter, such that the parameters to be inferred from data are only the regression coefficients (i.e., $\Theta_{\mathcal{C}} = \{\textbf{w}_{\mathcal{C}}\}$). As the prior is assumed to be uninformative, we avoid imposing any specific assumptions or biases on the parameter estimate. Accordingly, we set the conjugate prior as a zero-mean isotropic Gaussian distribution with infinitely broad variance, that is, $p(\textbf{w}_{\mathcal{C}} \vert \xi) = \mathcal{N}(\textbf{0},  \gamma^{-1} \textbf{I})$ with $\gamma \mapsto 0$ and $\textbf{I} \in \mathbb{R}^{\vert \mathcal{C} \vert \times \vert \mathcal{C} \vert}$ is the identity matrix. Let $\textbf{m}_{\mathcal{C}, t} \in \mathbb{R}^{\vert \mathcal{C} \vert}$ and $\textbf{K}_{\mathcal{C}, t} \in \mathbb{R}^{\vert \mathcal{C} \vert \times \vert \mathcal{C} \vert}$ denote the mean vector and the covariance matrix of the posterior at a particular time step $t$, respectively. As the posterior is Gaussian, we indeed know that its mode coincides with its mean, and since we can write the logarithm of the posterior as the sum of both the logarithm of the likelihood and the logarithm of the prior, the posterior mean reduces to the maximum likelihood estimate, given by
\begin{subequations}
    \begin{align}
        \textbf{m}_{\mathcal{C}, t}  &= \argmax_{\textbf{w}_{\mathcal{C}}} \ \log \left( p(\mathcal{D}_{\mathcal{C}, t} \vert \textbf{w}_{\mathcal{C}}, \xi) \right) + \log \left( p(\textbf{w}_{\mathcal{C}} \vert \xi) \right), \quad \forall t, \\
        &= \argmin_{\textbf{w}_{\mathcal{C}}} \ \sum_{t^\prime \leq t} ( y_{t^\prime} - f(\textbf{x}_{\mathcal{C}, t^\prime}, \textbf{w}_{\mathcal{C}}) )^2, \quad \forall t,
        \label{eq:maximum_likelihood}
    \end{align}
\end{subequations}
as would the estimated noise precision if inferred from data. Now suppose the data for one or more of the features is reported untruthfully, such that for any particular time step $t$, the vector of features is, $\tilde{\textbf{x}}_t = \textbf{x}_t + \boldsymbol{\eta}_t$, where $p(\boldsymbol{\eta}_{t}) = \mathcal{N}(\boldsymbol{0}, \boldsymbol{\Sigma})$. The covariance matrix $\boldsymbol{\Sigma}$ is diagonal, the elements of which corresponding to truthfully reported features are zero, such that only untruthful features are subject to noise.
For brevity, and without loss of generality, we consider only the set of linear basis functions. 
Substituting the new feature vector into the linear interpolant, the optimization problem in (\ref{eq:maximum_likelihood}) is augmented to minimize the expected sum-of-squares error, where the expectation is taken over the random noise,
\begin{align*}
     \textbf{m}_{\mathcal{C}, t} &= \argmin_{\textbf{w}_{\mathcal{C}}} \ \mathbb{E} \, \left[ \sum_{t^\prime \leq t} \left( y_{t^\prime} - f(\tilde{\textbf{x}}_{\mathcal{C}, t^\prime}, \textbf{w}_{\mathcal{C}}) \right)^2 \right], \quad \forall t, \\
     &= \argmin_{\textbf{w}_{\mathcal{C}}} \ \sum_{t^\prime \leq t} \left( ( y_{t^\prime} - \textbf{w}_{\mathcal{C}}^\top \textbf{x}_{\mathcal{C}, t^\prime} )^2 - 2 \textbf{w}_{\mathcal{C}}^\top \mathbb{E}  \left[ \boldsymbol{\eta}_{\mathcal{C}, t^\prime} \right] (y_{t^\prime} - \textbf{w}_{\mathcal{C}}^\top \textbf{x}_{\mathcal{C}, t^\prime}) +  \textbf{w}_{\mathcal{C}}^\top \mathbb{E} \left[ \boldsymbol{\eta}_{\mathcal{C}, t^\prime} \boldsymbol{\eta}_{\mathcal{C}, t^\prime}^\top \right] \textbf{w}_{\mathcal{C}} \right), \quad \forall t,      
     \\
     &= \argmin_{\textbf{w}_{\mathcal{C}}} \ \sum_{t^\prime \leq t} \left( y_{t^\prime} - \textbf{w}_{\mathcal{C}}^\top \textbf{x}_{\mathcal{C}, t} \right)^2 + \textbf{w}_{\mathcal{C}}^\top \boldsymbol{\Sigma} \textbf{w}_{\mathcal{C}}, \quad \forall t,
\end{align*}
where the last line is derived by recalling that $\mathbb{E}[\boldsymbol{\eta}_t] = \boldsymbol{0}$ and hence $\mathbb{E}[\boldsymbol{\eta}_t \boldsymbol{\eta}_t^\top] = \boldsymbol{\Sigma}$. For the special case whereby equal noise is added to all features, such that $\boldsymbol{\Sigma} = \beta \mathbf{I}$, for some constant $\beta$, we have  that $\textbf{w}_{\mathcal{C}}^\top \boldsymbol{\Sigma} \textbf{w}_{\mathcal{C}} = \beta \vert \vert \textbf{w}_{\mathcal{C}} \vert \vert^2_2$, resulting in a vector of coefficients analogous to that obtained using Ridge regression \citep{hoerl1970ridge}, with shrinkage penalty $\beta = \gamma / \xi$. Moreover, since $\boldsymbol{\Sigma}$ is a diagonal matrix, agents are not able to behave spitefully by adding noise to their features in effort to reduce the payments of others. In general though, we observe that the likelihood is maximized as $\gamma \mapsto 0$ and any addition of noise will create an endogeneity bias and reduce the predictive likelihood and subsequently the revenues, given the definition in (\ref{eq:seller_payment}), thereby completing the proof. 
\hfill\BlackBox

\section{Proof of Corollary~\ref{cor:kld_marginal_contribution_asymptotic}} \label{app:kld_marginal_contribution_asymptotic}
\begin{sublemma} \label{lem:expected_squared_diff} 
The difference between the expected value of the quadratic loss of two maximum likelihood linear regression models, both with a Gaussian likelihood function, is equivalent to the expected squared difference in their interpolated functions.
\end{sublemma}
\noindent {\bf Proof} [Lemma~\ref{lem:expected_squared_diff}] In a maximum likelihood linear regression setting with a Gaussian likelihood, the objective is proportional to the expected value of the quadratic loss which we can decompose as follows:
\begin{subequations}
    \begin{align}
        \mathbb{E} \, \left[ \left( f_{\mathcal{C}, t} - y_t \right)^2 \right] &= var \left( f_{\mathcal{C}, t} - y_t \right) + \mathbb{E} \left[ f_{\mathcal{C}, t}  - y_t \right]^2, \\
        &= var \left( f_{\mathcal{C}, t} \right) + var \left( y_t \right) - 2 \, cov \left( f_{\mathcal{C}, t}, y_t \right), \\
        &= var \left( y_t \right) - var \left( f_{\mathcal{C}, t} \right).
        \label{eq:variance_of_prediction}
    \end{align}
\end{subequations}
where for brevity, we write $f_{\mathcal{C}, t} = f(\mathbf{x}_{\mathcal{C}, t}, \mathbf{w}_{\mathcal{C}}), \ \forall t$. We can write the expression in (\ref{eq:variance_of_prediction}) since the covariance between the prediction and the target is equal to the variance of the prediction itself since the estimators are unbiased.
If we let $\mathcal{C}_{i} = \mathcal{C} \cup \{i\}$ denote the addition of the $i$-th feature index to a particular coalition for all $C \subseteq \mathcal{I}_{-c}$, we get that
\begin{subequations}
    \begin{align}
        \mathbb{E} \, \left[ \left( f_{\mathcal{C}, t} - y_t \right)^2 \right] - \mathbb{E} \, \left[ \left( f_{\mathcal{C}_i, t} - y_t \right)^2 \right] &= var \left( f_{\mathcal{C}_i, t} \right) - var \left( f_{\mathcal{C}, t} \right), \\
        &= \mathbb{E} \left[ \left( f_{\mathcal{C}_i, t} \right)^2 \right] - \mathbb{E} \left[ f_{\mathcal{C}_i, t}\right]^2 - \mathbb{E} \left[ \left( f_{\mathcal{C}, t}\right)^2 \right] + \mathbb{E} \left[ f_{\mathcal{C}, t}\right]^2, \\
        &= \mathbb{E} \left[ \left( f_{\mathcal{C}_i, t}\right)^2 \right] - \mathbb{E} \left[ \left( f_{\mathcal{C}, t}\right)^2 \right], \\
        &= \mathbb{E} \left[ \left( f_{\mathcal{C}_i, t}\right)^2 \right] + \mathbb{E} \left[ \left( f_{\mathcal{C}, t}\right)^2 \right] - 2  \mathbb{E} \left[ \left( f_{\mathcal{C}, t}\right)^2 \right], \\
        &= \mathbb{E} \left[ \left( f_{\mathcal{C}_i, t}\right)^2 \right] + \mathbb{E} \left[ \left( f_{\mathcal{C}, t}\right)^2 \right] - 2  \left( var \left( f_{\mathcal{C}_i, t} \right) + \mathbb{E} \left[ f_{\mathcal{C}_i, t} \right]^2 \right) \label{eq:expected_squared_diff}.
    \end{align}
\end{subequations}

Note that we can ignore the covariance $cov \left( f_{\mathcal{C}, t}, \left( f_{\mathcal{C}_i, t} - f_{\mathcal{C}, t} \right )\right)$ since the prediction is not correlated
with the residuals. Hence, the variance term can be re-written as follows:
\begin{align*}
    var \left( f_{\mathcal{C}_i, t} \right) 
    &= var \left( f_{\mathcal{C}, t} + \left( f_{\mathcal{C}_i, t} - f_{\mathcal{C}, t} \right) \right), \\
    &= var \left( f_{\mathcal{C}, t} \right) + var \left( f_{\mathcal{C}_i, t} - f_{\mathcal{C}, t} \right), \\
    &= 2 var \left( f_{\mathcal{C}, t} \right) + var \left( f_{\mathcal{C}_i, t} \right) - 2  cov \left( f_{\mathcal{C}_i, t}, f_{\mathcal{C}, t} \right),
\end{align*}
and we hence get that $var \left( f_{\mathcal{C}, t} \right) = cov \left( f_{\mathcal{C}_i, t}, f_{\mathcal{C}, t} \right)$. 
Given that upon standardization, $\mathbb{E} \left[ f_{\mathcal{C}, t} \right] = 0$ and $\mathbb{E} \left[ f_{\mathcal{C}_i, t} \right] = 0$, we can re-write the last term in (\ref{eq:expected_squared_diff}) as follows:
\begin{align*}
    var \left( f_{\mathcal{C}_i, t} \right) + \mathbb{E} \left[ f_{\mathcal{C}_i, t} \right]^2 &= cov \left( f_{\mathcal{C}_i, t}, f_{\mathcal{C}} \right) + \mathbb{E} \left[ f_{\mathcal{C}_i, t} \right]  \mathbb{E} \left[ f_{\mathcal{C}, t} \right], \\
    &= \mathbb{E} \left[ f_{\mathcal{C}_i, t} f_{\mathcal{C}, t} \right].
\end{align*}

Therefore, the difference in expected values of the quadratic loss reduces to the following:
\begin{align*}
    \mathbb{E} \, \left[ \left( f_{\mathcal{C}, t} - y_t \right)^2 \right] - \mathbb{E} \, \left[ \left( f_{\mathcal{C}_i, t} - y_t \right)^2 \right] &= \mathbb{E} \left[ \left( f_{\mathcal{C}_i, t} \right)^2 \right] + \mathbb{E} \left[ \left( f_{\mathcal{C}, t} \right)^2 \right] - 2 \mathbb{E} \left[ f_{\mathcal{C}_i, t} f_{\mathcal{C}, t} \right], \\
    &= \mathbb{E} \left[ \left( f_{\mathcal{C}_i, t} - f_{\mathcal{C}, t} \right)^2\right],
\end{align*}
which completes the proof. \hfill\BlackBox

\begin{sublemma} \label{lem:kl_definition}
Under Assumption~\ref{as:gaussian_hypothesis}, the expected KL divergence between two predictive distributions is asymptotically equivalent to the expected difference in their predictive means.
\end{sublemma}

\noindent {\bf Proof} [Lemma~\ref{lem:kl_definition}]. Following the notation as in Appendix~\ref{app:truthfulness}, we can write the general expression for the posterior predictive distribution in (\ref{eq:predictive_distribution}) as
\begin{align}
    p(y_{t} \, \vert \, \textbf{x}_{\mathcal{C}, t}) = \mathcal{N}( f(\textbf{x}_{\mathcal{C}, t}, \textbf{m}_{\mathcal{C}, t}), \xi_{\mathcal{C}, t}), \quad \forall t,
    \label{eq:gaussian_predictive_distribution}
\end{align}
where $\xi_{\mathcal{C}, t}$ is the precision (i.e., inverse variance) comprising the finite precision of the intrinsic noise and the uncertainty the coefficients such that, $1/\xi_{\mathcal{C}, t} = 1/\xi + \textbf{x}_{\mathcal{C}, t}^{\top} \textbf{K}_{\mathcal{C}, t} \textbf{x}_{\mathcal{C}, t}, \ \forall t$.

Let $Z_t = Y_t - f_{\mathcal{C}_i, t}, \, \forall t$. Since the predictive distribution is a univariate Gaussian, the logarithm of the likelihood ratio can be written as follows:
\begin{subequations}
    \begin{align}
         \log \left( \frac{p(y_{t} \, \vert \, \textbf{x}_{\mathcal{C}_i, t})}{p(y_{t} \, \vert \, \textbf{x}_{\mathcal{C}, t})} \right) 
         &= \frac{1}{2} \left( \frac{\frac{\sqrt{\xi_{\mathcal{C}, t}}} {\sqrt{2 \pi}}  \exp \left(\frac{1}{2} \xi_{\mathcal{C}, t} (Y_t - f_{\mathcal{C}, t})^2 \right) }{\frac{\sqrt{\xi_{\mathcal{C}_i, t}}} {\sqrt{2 \pi}}  \exp \left(\frac{1}{2} \xi_{\mathcal{C}_i, t} (Y_t - f_{\mathcal{C}_i, t})^2 \right)}  \right) \\
         &= \frac{1}{2} \left( \log \left( \frac{\xi_{\mathcal{C}_i, t}}{\xi_{\mathcal{C}, t}} \right) - \xi_{\mathcal{C}_i, t} (Y_t - f_{\mathcal{C}_i, t} )^2  - \xi_{\mathcal{C}, t} (Y_t - f_{\mathcal{C}, t} )^2 \right), \\
         &= \frac{1}{2} \log \left( \frac{\xi_{\mathcal{C}_i, t}}{\xi_{\mathcal{C}, t}} \right) - \frac{1}{2} ( \xi_{\mathcal{C}_i, t} - \xi_{\mathcal{C}, t} ) Z_t^2  -\xi_{\mathcal{C}, t} \Delta_{f_{\mathcal{C}_i, t}} Z_t  + \frac{1}{2} \xi_{\mathcal{C}, t} \left( \Delta_{f_{\mathcal{C}_i, t}} \right)^2,
         \label{eq:log_likelihood_ratio}
    \end{align}
\end{subequations}
where the expression, $\Delta_{f_{\mathcal{C}_i, t}} = f_{\mathcal{C}, t} - f_{\mathcal{C}_i, t}, \ \forall t$, denotes the difference in means.

The KL divergence is given by the expectation of (\ref{eq:log_likelihood_ratio}) with respect to the predictive distribution that includes the $i$-th feature. Whilst the expression is not linear in $Z_t$, recall that in general $\mathbb{E} [Z_t^2] = var \left[Z_t\right] + \mathbb{E}[Z_t]^2$. The variance and expected value of $Z_t$ reduces to $var \left[Z_t\right] =  var \left[ Y_t - f_{\mathcal{C}, t} \right] = 1 / \xi_{\mathcal{C}, t}$ and $\mathbb{E} [Z_t] = \mathbb{E} [Y_t - f_{\mathcal{C}_i, t}] = 0$, respectively. 

Given (\ref{eq:consistency}), we get that $\xi_{\mathcal{C}_i, t} \xrightarrow{\enskip t\enskip} \xi_{\mathcal{C}, t}$, hence we can derive the KL divergence as follows:
\begin{align*}
     \mathbb{E} \left[ D_{\textrm{KL}} \left(  p(y_{t} \vert \textbf{x}_{\mathcal{C}_i, t}) \vert\vert p(y_{t} \vert \textbf{x}_{\mathcal{C}, t})\right)\right] 
     &= \mathbb{E} \, \left[ \int_{Y_t}  p(y_{t} \, \vert \, \textbf{x}_{\mathcal{C}_i, t}) \, \log \left\{ \frac{p(y_{t} \, \vert \, \textbf{x}_{\mathcal{C}_i, t})}{p(y_{t} \, \vert \, \textbf{x}_{\mathcal{C}, t})} \right\} d {y_t}\right], \\
     &= \mathbb{E} \, \left[ \frac{1}{2} \left( \log \left\{ \frac{\xi_{\mathcal{C}_i, t}}{\xi_{\mathcal{C}, t}} \right\} - 1 + \frac{\xi_{\mathcal{C}, t}}{\xi_{\mathcal{C}_i, t}} + \xi_{\mathcal{C}, t} \Delta_{f_{\mathcal{C}_i, t}}^2 \right) \right],  \\
     &\xrightarrow{\enskip t\enskip} \frac{\xi}{2} \, \mathbb{E} \, \left[ \left( f_{\mathcal{C}_i, t} - f_{\mathcal{C}, t} \right)^2 \right], \ \mathrm{almost} \ \mathrm{surely},
\end{align*}
which completes the proof \hfill\BlackBox

\noindent {\bf Proof} [Corollary~\ref{cor:kld_marginal_contribution_asymptotic}]. The marginal contribution derived using the expected NLL is equivalent to that obtained using the quadratic loss, and hence given by an in-sample estimate of the following, 
\begin{align}
     m_{i, t}(\mathcal{C}) = \frac{\xi^2}{2} \left( \mathbb{E} \, \left[ \left( f_{\mathcal{C}, t} - y_t \right)^2 \right] - \mathbb{E} \, \left[ \left( f_{\mathcal{C}_i, t} - y_t \right)^2 \right] \right).
     \label{eq:mse_contribution}
\end{align}

Combining the results from Lemmas~\ref{lem:expected_squared_diff} and \ref{lem:kl_definition}, we can see that $(\ref{eq:kld_marginal_contribution}) \xrightarrow{\enskip t\enskip} (\ref{eq:mse_contribution})$, and therefore, since all other terms within the definition in (\ref{eq:shapley_value}) remain unchanged, the Shapley values, and therefore the payments, will converge, thereby completing the proof.
\hfill\BlackBox

\section{Proof of Budget Balance Violation (Theorem~\ref{the:kld_marginal_contribution})} \label{app:budget_balance_violation}
We provide here a proof that budget balance is violated under the definition of marginal contribution in (\ref{eq:kld_marginal_contribution}), as described in Theorem~\ref{the:kld_characteristic_function_asymptotic}.

Recall that budget balance proceeds from the semivalue axiom \textit{efficiency}, which in our context translates to: total attribution allocated to all features should sum to the value of the grand coalition, that is, $v_t (\mathcal{I}_c) - v_t (\mathcal{I}) = \sum_{i \in \mathcal{I}_{-c}} \phi_{i, t}, \ \forall t$. Using (\ref{eq:shapley_value}) we can expand this definition thereby revealing the telescoping sum structure of the Shapley value such that
\begin{subequations}
    \begin{align}
        \sum_{i \in \mathcal{I}_{-c}} \phi_{i, t} =& \sum_{i \in \mathcal{I}_{-c}} \sum_{\mathcal{C} \in 
        \mathcal{P}(\mathcal{I}_{-c} \setminus \{i\})
        } \frac{\vert\mathcal{C}\vert!(\vert\mathcal{I}_{-c}\vert - \vert\mathcal{C}\vert - 1)!}{\vert\mathcal{I}_{-c}\vert!} \, \left( \mathcal{C} \cup \mathcal{I}_c) - v_t (\mathcal{C} \cup \mathcal{I}_c \cup \{i\}) \right), \quad \forall t, \label{eq:telescoping_sum} \\
        =& \nonumber \vert \mathcal{I}_{-c} \vert \underbracket{\frac{0! ( \vert \mathcal{I}_{-c} \vert - 1)!}{\vert \mathcal{I}_{-c} \vert!} v_t (\mathcal{I}_c)}_{\substack{\textrm{The value of the \textbf{central}}\\\textrm{\textbf{agent} appears $|\mathcal{I}_{-c}|$}\\\textrm{times.}}} - \quad \vert \mathcal{I}_{-c} \vert \underbracket{\frac{( \vert \mathcal{I}_{-c} \vert - 1)! 1!}{\vert \mathcal{I}_{-c} \vert!} v_t (\mathcal{I})}_{\substack{\textrm{The value of the \textbf{grand}}\\\textrm{\textbf{coalition} appears $|\mathcal{I}_{-c}|$}\\\textrm{times.}}} \\
        & +  \sum_{\substack{\mathcal{C} \in  \mathcal{P}(\mathcal{I}_{-c})\\\ \mathcal{C} \neq \emptyset}} (\vert \mathcal{I}_{-c} \vert - \vert \mathcal{C} \vert) \underbracket{\left( \frac{\vert \mathcal{C} \vert ! (\vert \mathcal{I}_{-c} \vert - \vert \mathcal{C} \vert - 1) !}{\vert \mathcal{I}_{-c} \vert !} \right) v_t (\mathcal{C} \cup \mathcal{I}_c)}_{\substack{\textrm{The value of the coalition $\mathcal{C}$ appears}\\\textrm{$|\mathcal{I}_{-c}| - |\mathcal{C}|$ times with a \textbf{positive}}\\\textrm{sign, i.e., once per agent \textbf{in} $\mathcal{C}$.}}}  \\
        & - \nonumber \sum_{\substack{\mathcal{C} \in  \mathcal{P}(\mathcal{I}_{-c})\\\ \mathcal{C} \neq \emptyset}} \vert \mathcal{C} \vert \underbracket{\left( \frac{(\vert \mathcal{C} \vert - 1)! (\vert \mathcal{I}_{-c} \vert - \vert \mathcal{C} \vert) !}{\vert \mathcal{I}_{-c} \vert !} \right) v_t (\mathcal{C} \cup \mathcal{I}_c)}_{\substack{\textrm{The value of the coalition $\mathcal{C}$ appears}\\\textrm{$|\mathcal{C}|$ times with a \textbf{negative} sign, i.e.,}\\\textrm{once per agent \textbf{not in} $\mathcal{C}$.}}}, \quad \forall t, \\
        =& v_t (\mathcal{I}_) - v_t (\mathcal{I}), \quad \forall t.
    \end{align}
\end{subequations}

However, if we replace the expression for the marginal contribution with the definition in (\ref{eq:kld_marginal_contribution}), we can similarly derive an expression for the value of the grand coalition, given by
\begin{align*}
    \sum_{i \in \mathcal{I}_{-c}} \phi_{i, t} = \vert \mathcal{I}_{-c} \vert \frac{(\vert \mathcal{I}_{-c} \vert - 1)!}{\vert \mathcal{I}_{-c} \vert !} \, \mathbb{E} \left[ D_{\textrm{KL}} \left( p(y_{t} \vert  \textbf{x}_{\mathcal{I}, t}, \mathcal{D}_{\mathcal{I}, t-1}) \vert \vert p(y_{t} \vert  \textbf{x}_{\mathcal{I}_c , t}, \mathcal{D}_{\mathcal{I}_c, t-1}) \right) \right].
\end{align*}

This, however, holds true if and only if the KL divergence satisfies the triangle inequality with equality. Yet, we know that statistical divergence metrics do not satisfy the triangle inequality, that is, for any probability densities, $X$, $Y$ and $Z$, we get that $D_{\textrm{KL}} (X \vert\vert Z) \not \leq D_{\textrm{KL}} ( X \vert\vert Y) + D_{\textrm{KL}} (Y \vert \vert Z )$. Hence, by contradiction we prove that using the KL divergence in this manner violates the efficiency axiom, and subsequently budget balance cannot be guaranteed.
\hfill\BlackBox

\section{Proof of Corollary~\ref{cor:kld_characteristic_function_asymptotic}} \label{app:kld_characteristic_function_asymptotic}

\noindent {\bf Proof} [Corollary~\ref{cor:kld_characteristic_function_asymptotic}].
In (\ref{eq:kld_characteristic_function}), we defined the valuation of a coalition incorporating the KL divergence as below, which in Appendix~\ref{app:kld_marginal_contribution_asymptotic}, was shown to converge to
\begin{align*}
    v_t(\mathcal{C}) &= \mathbb{E} \left[ D_{\textrm{KL}} \left( p(y_{t} \vert \textbf{x}_{\mathcal{C}, t}) \vert\vert p(y_{t} \vert \textbf{x}_{\mathcal{I}_c, t}) \right) \right], \\
    &\xrightarrow{\enskip t\enskip} \frac{\xi^2}{2} \left( \mathbb{E} \, \left[ \left( f_{\mathcal{I}_c, t} - y_t \right)^2 \right] - \mathbb{E} \, \left[ \left( f_{\mathcal{C}, t} - y_t \right)^2 \right] \right), \, \mathrm{almost \ surely},
\end{align*}
therefore the marginal contribution of a feature to a coalition converges to the following:
\begin{subequations}
    \begin{align}
         m_{i, t}(\mathcal{C}) &\xrightarrow{\enskip t\enskip} \frac{\xi^2}{2} \left( \mathbb{E} \, \left[ \left( f_{\mathcal{I}_c, t} - y_t \right)^2 \right] - \mathbb{E} \, \left[ \left( f_{\mathcal{C}_i, t} - y_t \right)^2 \right]  -  \mathbb{E} \, \left[ \left( f_{\mathcal{I}_c, t} - y_t \right)^2 \right] - \mathbb{E} \, \left[ \left( f_{\mathcal{C}, t} - y_t \right)^2 \right] \right), \\
         &= \frac{\xi^2}{2} \left( \mathbb{E} \, \left[ \left( f_{\mathcal{C}, t} - y_t \right)^2 \right] - \mathbb{E} \, \left[ \left( f_{\mathcal{C}_i, t} - y_t \right)^2 \right] \right), \\
         &= (\ref{eq:mse_contribution}).
    \end{align}
\end{subequations}
As in Appendix~\ref{app:kld_marginal_contribution_asymptotic}, since all other terms within the definition in (\ref{eq:shapley_value}) remain unchanged, the Shapley values, and therefore the payments, will converge, thereby completing the proof.
\hfill\BlackBox

\vskip 0.2in
\bibliography{references}

\end{document}